\documentclass[lettersize,journal]{IEEEtran}
\usepackage{amsmath,amsfonts}
\usepackage{algorithmic}
\usepackage{algorithm}
\usepackage{array}
\usepackage[caption=false,font=normalsize,labelfont=sf,textfont=sf]{subfig}
\usepackage{textcomp}
\usepackage{stfloats}
\usepackage{url}
\usepackage{verbatim}
\usepackage{graphicx}
\usepackage{cite}
\usepackage{xcolor}
\usepackage{booktabs}
\usepackage{multirow} 
\usepackage{threeparttable}
\begin{document}
\title{NLP-enabled Trajectory Map-matching in Urban Road Networks using a Transformer-based Encoder-decoder}
\author{Sevin Mohammadi$^{1}$,  and Andrew W. Smyth$^*{^{2}}$
\thanks{$^{1}$Sevin Mohammadi is a Ph.D. candidate in the Civil Engineering and Engineering Mechanics Department at Columbia University,
        New York, NY 10027, USA
        {\tt\small sm4894@columbia.edu}}%
\thanks{$^{2}$Andrew W. Smyth (corresponding author) is the Robert A. W. and Christine S. Carleton Professor of Civil Engineering and Engineering Mechanics and the Director of the NSF Center for the Smart Streetscapes (CS3) at Columbia University,
        New York, NY 10027, USA
        {\tt\small aws16@columbia.edu}}%
}

\markboth{Journal of \LaTeX\ Class Files,~Vol.~14, No.~8, August~2021}%
{Shell \MakeLowercase{\textit{et al.}}: A Sample Article Using IEEEtran.cls for IEEE Journals}


\maketitle

\begin{abstract}
Vehicular trajectory data, collected through geolocation telematics, is crucial for understanding mobility patterns in urban road networks. Map-matching, the process of aligning noisy and sparsely sampled GPS trajectories with digital road network map, is key to reconstructing accurate vehicle trajectories. Traditional methods rely on geometric proximity, topology, and shortest-path heuristics, overlooking two critical factors: (1) drivers often favor routes based on local road characteristics rather than shortest paths, revealing shared preferences that can be learned from large-scale trajectory data, and (2) GPS noise varies spatially mainly due to spatial variations in multipath effects. The interplay between these factors may make conventional methods less effective in complex real-world scenarios and may be more time-consuming for human experts to identify such heuristics and implement them. This study introduces an end-to-end and fully data-driven deep learning-based offline map-matching framework that formulates trajectory map-matching as a machine translation task, inspired by natural language processing (NLP). Specifically, it presents a transformer-based encoder-decoder map-matching framework that automatically learns contextual representations of noisy GPS points to infer drivers' trajectory behavior and road network structures in an end-to-end manner. Trained on large-scale trajectory data, the proposed method improves path estimation accuracy in map-matching. Experiments on synthesized trajectories demonstrate that the proposed transformer-based map-matching framework outperforms conventional methods by integrating contextual awareness into trajectory analysis. Further evaluation on real-world GPS traces from Manhattan, New York, shows that our model achieves 75\% accuracy in reconstructing navigated routes. These results demonstrate the advantage of transformer-based architectures in capturing real-world drivers' trajectory behaviors, complex spatial dependencies and noise patterns, providing a more robust and scalable context-aware solution for map-matching. This study contributes to the broader vision of advancing trajectory-driven foundation models to enhance geospatial modeling and accelerate downstream urban mobility applications.
\end{abstract}

\begin{IEEEkeywords}
Trajectory analysis, telematics data, GPS, GNSS, map-matching, transformer, sequential data modeling, encoder-decoder, deep Learning.
\end{IEEEkeywords}

\section{INTRODUCTION}
The emergence of smart cities and Intelligent Transportation Systems (ITS) marks a significant advancement in urban mobility planning and management \cite{batty2012smart}. Leveraging advanced technologies, these systems propel the development of transportation infrastructures \cite{gao2021digital}, improve traffic safety \cite{abdel2023real,khattak2021big,alrassy2023driver}, and contribute to the growth of smart, sustainable cities \cite{bibri2017smart}. Central to this paradigm are Connected Vehicles (CVs), which enable vehicle-to-everything (V2X) communication including Vehicle-to-Vehicle (V2V) and Vehicle-to-Infrastructure (V2I) interactions, creating a network for intelligent urban navigation \cite{lu2014connected}. The ubiquity of the communication technologies has facilitated the collection of large amounts of real-world data, ranging from vehicle locations and speeds to traffic flow patterns. Analyzing this data offers insights into traffic patterns and drivers' driving behavior, improving traffic modeling \cite{mohammadi2021role} and safety \cite{alrassy2023driver}.
Among the pool of data gathered by connected vehicles, vehicular trajectory data holds numerous applications encompassing monitoring transportation systems and urban infrastructure, safety management \cite{liu2017comparative,li2021using,alrassy2023driver}, public transit efficiency improvement \cite{pinelli2016data,nguyen2018angeles}, ride-hailing services \cite{hui2021trajnet}, travel time prediction \cite{mohammadi2023probabilistic,olivier2023bayesian}, and optimizing emergency response systems \cite{olivier2022data,de2021simulating}. By tracking the movement of vehicles, trajectory data facilitates route optimization and the identification of congestion zones and enhances the efficiency of urban road networks. Markovi et al. review trajectory data applications in different domains such as modeling human behavior, designing public transit, traffic performance measurement and prediction, environment, and safety \cite{markovic2018applications}. 

\begin{figure}[ht]
      \centering
      \includegraphics[scale=0.38]{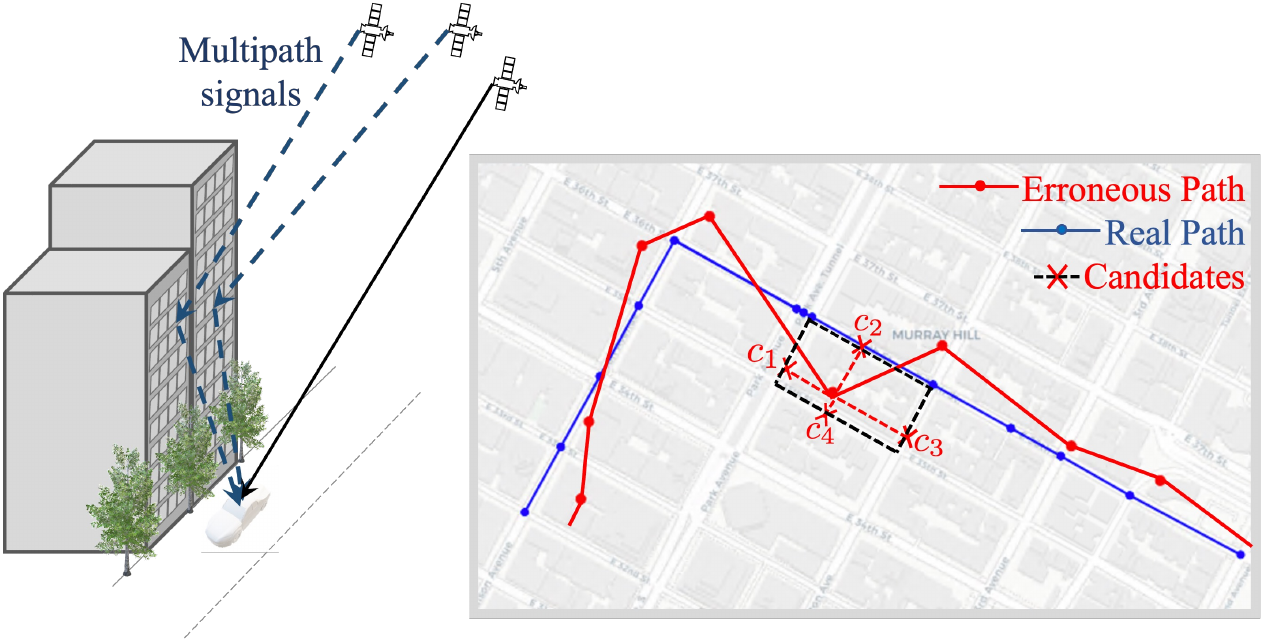}
      \caption{Left: Illustration of signal reflections caused by tall buildings in an urban environment, leading to multipath errors. Right: Comparison between the erroneous path in red and the actual path in blue within Manhattan. Map-matching accurately estimates the true path. The illustration provides an example of all candidate segments, denoted as $C_1, C_2, C_3, C_4$, for one of the erroneous raw GPS points collected off the road segment.}
      \label{fig:error_illustration}
   \end{figure}

A vehicle trajectory represents the path a vehicle follows during a trip over time, typically expressed as a sequence of timestamped geolocation points. These points are collected by GPS receivers installed in the vehicle, which continuously record its position at specific time intervals. These receivers determine the vehicle's position by communicating with a network of satellites orbiting Earth. However, in urban environments, the presence of tall buildings creates an ``urban canyon" effect, which introduces errors in position calculations and consequently degrades the accuracy of GPS points. A major factor contributing to this issue is the multipath effect, where GPS signals reflect off surfaces such as buildings and reach the receiver via indirect paths. This causes incorrect position calculations due to the additional travel time of the reflected signals. Figure \ref{fig:error_illustration} shows an schematic example of the multipath effect and compares the resulting erroneous trajectory with the actual route on a map.
In addition to the multipath effect, signal blockage and limited satellite visibility further degrade the accuracy of GPS trajectory data in urban settings \cite{karaim2018gnss}. Moreover, the dense and narrow street layouts of urban areas make it even more challenging to reconstruct the true path from a sequence of noisy GPS points. This challenge is further compounded when the GPS sampling rate is low, leading to a trajectory represented by a sparse sequence of noisy data points. 

Therefore, a key challenge in utilizing trajectory data for downstream applications is the preprocessing of noisy and sparse GPS records. Map-matching techniques are used to align raw GPS points with the underlying road network, correcting positioning errors and inferring road segments between consecutive GPS points to eventually reconstruct the most likely vehicle trajectory. These techniques utilize a broad range of algorithms, from statistical to data-driven approaches, including geometric methods, probabilistic models, and advanced machine learning models \cite{quddus2007current,hashemi2014critical}.

A preliminary step in the majority of advanced map-matching algorithms involves the selection of candidate road segments for each observed GPS point (Figure \ref{fig:error_illustration} shows an example). Typically, road segments that fall within a search region, often represented as an ellipse or a circle centered around the GPS point, are chosen as candidate roads. The dimension of the search region is determined based on the positioning error statistics, which is either a fixed distance threshold or an adaptive region \cite{alrassy2020map}. The algorithm then determines the most likely path by iteratively evaluating the set of all possible routes formed from the candidate segments.
Path estimation from candidate road segments, i.e., identifying the most likely route, varies across map-matching algorithms, which apply geometric, topological, and probabilistic criteria. For instance, probabilistic models, such as Hidden Markov Models (HMMs), aim to maximize the joint emission and transition probabilities. Emission probability is the likelihood of observing noisy GPS points given that the vehicle is on a specific road segment. It is often calculated using the distance proximity and an estimate of the noise magnitude for the road network. Transition probability quantifies the conditional likelihood of transitioning from one road segment to the next. It is usually determined based on the difference between the observed, potentially erroneous trajectory and the shortest path between consecutive candidate points \cite{lou2009map}.

Path estimation in conventional map-matching algorithms primarily relies on geometric proximity, topology, and shortest-path heuristics, overlooking two critical factors that makes them less effective for map-matching sparse noisy data: (1) drivers often favor routes based on local road characteristics rather than shortest paths, revealing shared preferences that can be learned from large-scale trajectory data, and (2) GPS noise varies spatially mainly due to spatial variations in multipath effects. Street layouts and surrounding buildings contribute to the multipath effect, causing noise levels to fluctuate based on urban density. As illustrated in Figure 9, noise has a spatially non-uniform pattern which is high in dense urban areas with tall buildings and narrow streets. Most probabilistic map-matching approaches assume uniform GPS noise, making them noise sensitive. The interplay between drivers' trajectory behavior, road network structure, and spatially varying noise makes conventional methods less effective in the real-world where data is typically noisy and sparsely collected. Therefore, in path estimation in the trajectory map-matching, context-aware approaches \cite{choudhury2024towards}, are essential for effectively managing varying noise and drivers’ trajectory patterns.

Recent advancements in data-driven approaches, particularly deep learning and large-scale neural network models, have demonstrated remarkable success in learning contextual structures and capturing complex patterns across diverse domains, from drug discovery \cite{chithrananda2020chemberta} to computer vision and natural language processing by leveraging large domain-specific datasets. Similarly, in geospatial modeling, these models effectively capture intricate mobility patterns by learning relationships between road network structure, the built environment as a key driver of spatial noise variation, and drivers' navigation preferences across different locations. By autonomously learning structural features of the road network and uncovering hidden contextual factors influencing route choices and noise patterns, these models integrate this knowledge in an end-to-end manner to estimate the vehicular trajectory.

\begin{figure*}
 \centering
      \includegraphics[scale=0.66]{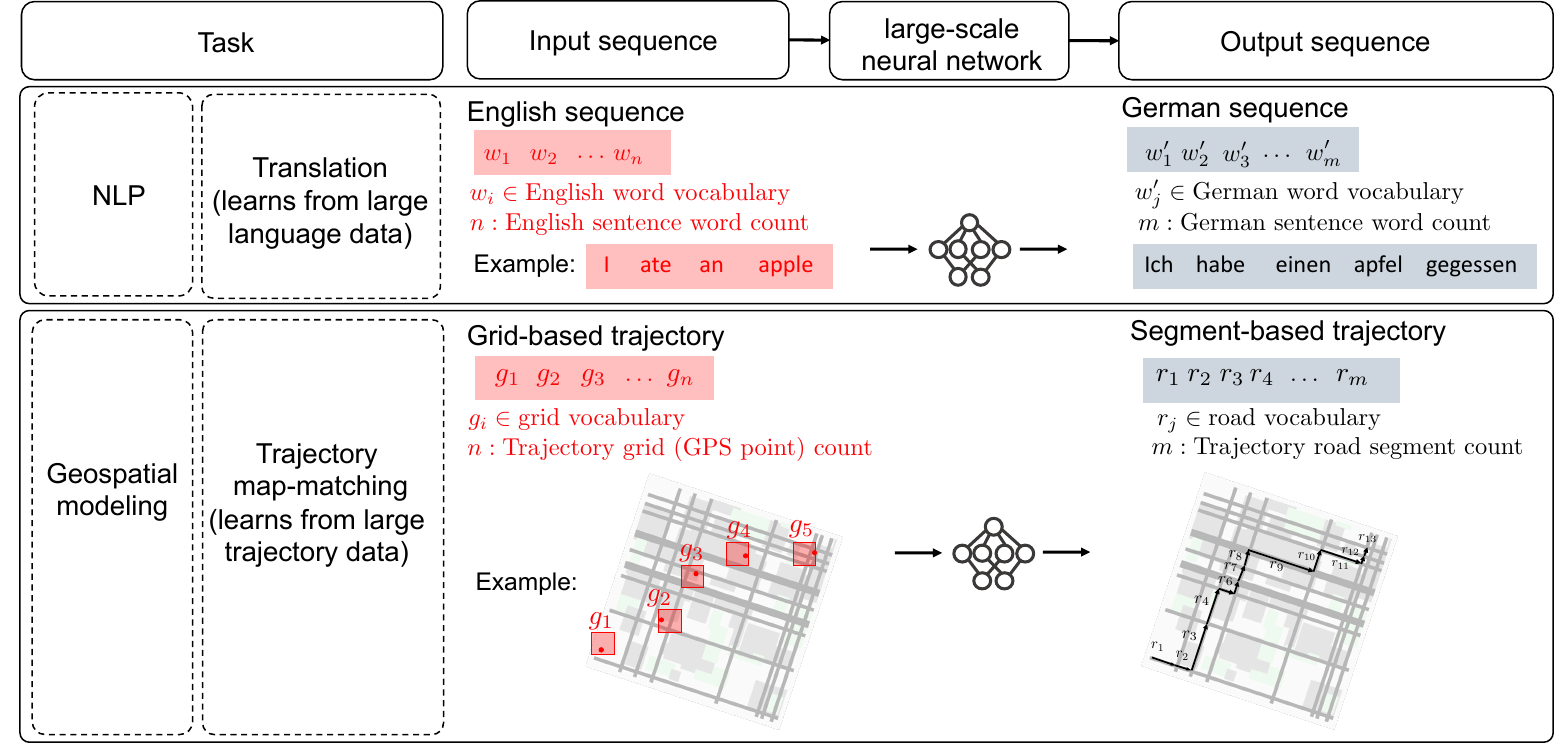}
      \caption{NLP-enabled trajectory map-matching: recovering a complete, ordered sequence of connected road segments from a noisy, sparse, and discretized sequence of GPS points.}
      \label{fig:nlp_analogy}   
\end{figure*}

In this study, we draw an analogy between trajectory map-matching and the translation task in natural language processing (NLP). We conceptualize vehicle trajectories within large spatial datasets as analogous to sentences within a text corpus. Just as NLP models learn a variety of contextual patterns from large-scale language data to perform downstream tasks such as translation, similar architectures can be leveraged to extract contextual factors relevant to vehicle trajectories from large trajectory datasets. Therefore, we frame trajectory map-matching as a machine translation problem. As Figure \ref{fig:nlp_analogy} shows, in the former, a sequence of noisy, sparse GPS points is translated to an ordered sequence of road segments, while in the latter, a sequence of words in one language is translated to a sequence of words in another language. A similar analogy is presented by \cite{musleh2024let}, defining map-matching as trajectory imputation by identifying its missing points, similar to filling in missing words in a sentence. Their work highlights the effectiveness of NLP models, particularly Bidirectional Encoder Representations from Transformers (BERT) \cite{kenton2019bert}, for the trajectory map-matching task.

In a similar fashion, this paper employs an encoder-decoder architecture with a transformer that leverages attention mechanisms \cite{vaswani2017attention} for the map-matching task. Trained on a large trajectory dataset, the encoder learns dense vector embeddings of tokenized, discretized GPS points (Figure \ref{fig:inputoutput} presents an example of discretized GPS points), capturing contextual structures such as spatial correlations. Meanwhile, the decoder learns embeddings of tokenized road segments, modeling the relationships between GPS points and road network features. Once trained, the model infers the sequence of road segments that best represents the traveled path given noisy, sparse GPS points. The rationale behind using the proposed deep learning sequence-to-sequence model is their capability to capture complex spatiotemporal dependencies, contextual factors influencing drivers’ trajectory behavior, and spatial noise variations across the road network by leveraging large-scale trajectory data.

Unlike traditional approaches such as Hidden Markov Models (HMMs) and other rule-based methods, transformers leverage embedding representations and attention mechanisms to learn relationships within and between GPS points and road network structures. This allows them to capture both local road network features and global contextual dependencies without relying on predefined transition probabilities or manually crafted rules. Consequently, transformers can adapt to varying noise levels across regions and account for drivers’ trajectory behaviors, enabling them to integrate information effectively when inferring missing road segments and accurately reconstructing the navigation path from sparse, noisy location points.

Moreover, large-scale neural network models, such as transformers, have enabled the development of pretrained foundation models across various domains, from natural language processing \cite{achiam2023gpt} to medicine \cite{moor2023foundation}. When trained on extensive trajectory datasets from cities with diverse road structures, mobility patterns, and GPS noise distributions, these models have the potential to evolve into scalable, region-agnostic tools for trajectory analysis. However, realizing this adaptability requires further investigation. Advancing in this direction aligns with the vision of a universal trajectory foundation model, which could automate urban mobility analytics and accelerate trajectory analysis for a broad range of downstream mobility and geospatial applications.

\section{Related Works}
In this section, a brief review of existing map-matching algorithms is presented. More comprehensive reviews of these algorithms are provided by \cite{quddus2007current,hashemi2014critical,kubicka2018comparative,chao2020survey}. Early map-matching algorithms were primarily search-based, focusing on the identification of the nearest road segment to a raw GPS point based solely on the geometry of the road segments \cite{white2000some,bernstein1996introduction}. This approach, known as geometric map-matching, disregarded the connectivity of the roads. However, with further advancements, topological matching algorithms were introduced, which took into account the contiguity of geometric shapes \cite{quddus2003general,greenfeld2002matching}. Building upon these developments, researchers then proposed probabilistic algorithms that relied on GPS error statistics to establish a confidence region \cite{zito1995global,ochieng2003map}. This error region enabled the selection of multiple candidate segments, each to be evaluated using different criteria.

These pioneering studies formed the basis for the subsequently advanced map-matching algorithms that utilize sophisticated techniques, such as Hidden Markov Models (HMM) \cite{hummel2006map,newson2009hidden,goh2012online,koller2015fast}, and weight-graph technique \cite{quddus2003general,lou2009map,li2013high,quddus2015shortest}, Kalman Filter \cite{el2005road,zhao2003extended}, Fuzzy logic models \cite{kim2001adaptive,quddus2006high}. 
Weight-based algorithms and HMMs are among the widely used algorithms. Weight-based map-matching algorithms typically assign scores to candidate road segment pairs, considering factors like spatial proximity, vehicle heading relative to segment direction, link connectivity, and turn restrictions at junctions \cite{quddus2015shortest}. These algorithms then select consecutive segments forming a least-cost path using methods like Dijkstra's shortest-path algorithm \cite{dijkstra2022note}. 
In HMM-based map-matching, where hidden states correspond to the road segments, the goal is to determine the most likely sequence of hidden states given a sequence of observations (here GPS points), typically accomplished through the use of the Viterbi algorithm \cite{forney1973viterbi}. HMMs model the sequence of states as a discrete-time Markov chain, where transitions to the next state depend solely on the current state. This reflects the Markov property that simplifies the modeling of dependencies between states. HMMs maximize the joint probability of the observed sequence and the hidden sequence by combining the emission probabilities and transition probabilities. The emission probability quantifies the likelihood of observing noisy GPS points given the vehicle's presence in a candidate segment. The transition probability represents the conditional likelihood of transitioning from one hidden state to another, equivalent to transitioning from one road segment to the next segment. It captures the dynamics of how the hidden states evolve over time in the Markov chain. Some map-matching algorithms employ a combination of weight-based models and HMMs that transforms the HMM problem into a Dijkstra least-cost path problem. They construct a virtual graph with candidate points as nodes and interconnecting paths as edges, assigning costs based on computed joint probabilities \cite{alrassy2020map,koller2015fast}.

However, a significant drawback of map-matching algorithms is their limited ability to effectively leverage historical data, rendering them vulnerable to noisy input. Deep learning-based map-matching models, in contrast, leverage large trajectory data during training which enables them to capture intricate patterns including noise and mobility within extensive trajectories, and thus makes them robust to noise. By learning complex representations from the data, these models effectively harness historical information, leading to significant improvements in map-matching performance while offering the advantage of efficient processing of large data. This combination of robustness, and efficient processing makes deep learning models valuable for map-matching task. 
The prevalent deep learning architecture for map-matching vehicle trajectories includes the sequence-to-sequence (seq2seq) models, which often incorporate an encoder-decoder structure including recurrent neural networks (RNNs) \cite{feng2020deepmm,ren2021mtrajrec,jiang2023l2mm} and graph neural networks  \cite{liu2023graphmm}. RNN-based seq2seq models are priorly used for vehicle trajectory prediction \cite{park2018sequence}, space-free vehicle trajectory recovery \cite{wang2019deep}, and human mobility prediction \cite{feng2018deepmove}. 
Later, Feng et al. utilized an RNN-based seq2seq model in particular LSTM-based encoder-decoder named DeepMM for map-matching. They enhanced their model's performance by applying data augmentation techniques that simulated the generation process of real-like raw trajectories. This approach allowed them to generate a substantial number of virtual trajectories from ground truth trajectories, thereby enriching their training dataset \cite{feng2020deepmm}. Although simulated trajectories enriches the training data, it overlooks the role of the underlying driving pattern and route choice behavior of the drivers. 
Ren et al. implemented a multi-task seq2seq learning architecture in particular GRU-based encoder-decoder named MTrajRec to predict the road segment and the moving ratios for each GPS point simultaneously \cite{ren2021mtrajrec}.
Moreover, Jian et al. harnessed a seq2seq GRU-based RNN model to refine the map-matching results for low-quality trajectory data by integrating data distribution learning from high-frequency trajectories and incorporating an explicit mobility pattern recognition module into their deep learning algorithm \cite{jiang2023l2mm}. This approach sets their work apart from other deep learning map-matching algorithms, which rely on the deep model's capacity to extract embedded mobility patterns from extensive trajectory data. 
In the context of seq2seq encoder-decoder models used for map-matching, RNNs have historically played a central role. However, there are inherent drawbacks associated with training RNN models. One well-known challenge is the vanishing or exploding gradient problem, which can hinder training effectiveness. Additionally, RNN-based encoder-decoder models incur high computational costs, particularly when processing lengthy sequences, making the processing of long sequences a limitation.

In this paper, we solve the map-matching task by constructing a seq2seq model based on the transformer architecture \cite{vaswani2017attention}. In this model, the RNN component is replaced with multiple self-attention modules in both the encoder and decoder, allowing the model to significantly outpace RNN-based encoder-decoder models in terms of accuracy and speed, particularly when processing a large set of lengthy GPS trajectories in urban network.

\section{Map-matching preliminaries}
This section outlines the definitions of the key components of the map-matching task and describes the data structure for the purpose of this paper. 

\textbf{Map-matching} is a technique used to align noisy GPS coordinates with road segments from a digital road network map. Thereby, map-matching reconstructs the most likely path taken by a vehicle as a sequence of connected road segments. This task becomes particularly challenging in urban environments due to the complex and dense road layout, compounded by significant GPS positioning errors caused by signal obstructions and multipath effects, as well as the sparsity of GPS points in the trajectory.
Map-matching tasks can be classified into two types: online and offline. Online map-matching algorithms process streaming GPS data in real time, providing immediate feedback on the moving vehicle's current position relative to the map. In contrast, offline map-matching algorithms analyze historical trajectory data after it has been fully collected, focusing on improving path estimation accuracy by leveraging the entire observed, albeit erroneous, trajectory. The processed trajectory data is then used for various downstream applications.

\textbf{Digital Map.} 
A digital map or digital road network map is a structured representation of roadway data that stores both geographical and topological information, along with associated attributes. It is typically represented as a directed graph, denoted as $Graph=(V,R)$, where $V$ represents a set of nodes corresponding to the endpoints of road segments, and $R$ represents a set of directed edges connecting these nodes. Each directed edge, $r\in R$, corresponds to a road segment with its direction indicating the allowed direction of traffic along that segment. A road segment is a component of a digital road map corresponding to a particular stretch of road in the actual road network. Each edge typically includes additional attributes, such as road type, and speed limit. 
A digital map can be obtained directly as a graph from sources such as OpenStreetMap or constructed from publicly available regional spatial tabular data. For example, the geographic base information of New York City streets, stored in the LION shapefile maintained by the NYC Department of City Planning \cite{LION}, can be used to generate a road network graph of the NYC.

\textbf{GPS-based Trajectory}. Raw GPS-based trajectory is a collection of timestamped, noisy location points recorded by a GPS device at a frequency determined by the device setting. The sampling rate may vary depending on the purpose of the data collection and the capacity of the data storage infrastructure, resulting in either low or high sampling trajectory data. The collected data points typically include the vehicle's position data, such as latitude, longitude, altitude, and speed, along with additional information, such as heading and the number of satellites the GPS device communicated with during location sampling.
For the map-matching task in this study, the raw GPS-based trajectory is defined as a sequence order of latitude and longitude pairs, denoted by $T_p= \left\{ p_1=(x_1, y_1), p_2=(x_2, y_2), \dots, p_n=(x_n, y_n) \right\}$. Here, $(x_i, y_i), i=1,\dots,n$ represents the geographic coordinates of the location points, and $n$ is the number of collected points along the route.

\begin{figure*}[thpb]
      \centering
      \includegraphics[scale=0.4]{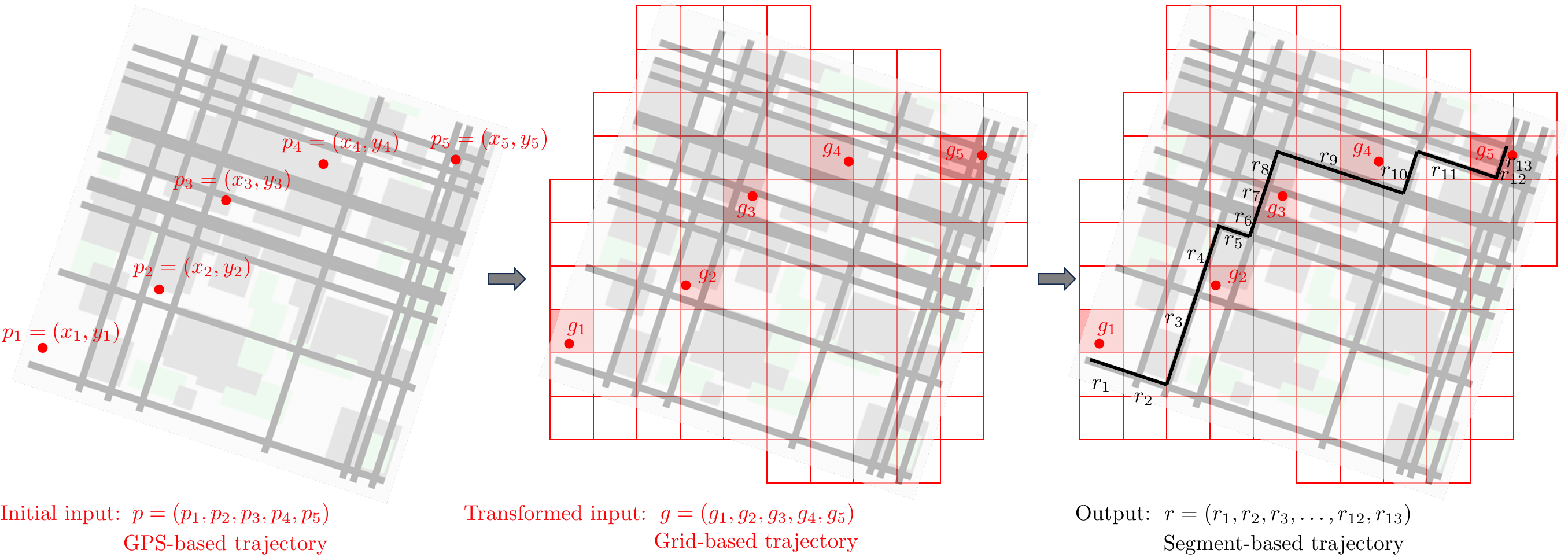}
      \caption{A schematic representation of the input and output of the map-matching model.}
      \label{fig:inputoutput}
   \end{figure*}

\textbf{Grid-based Trajectory}. In spatial data analysis, the spatial region is often divided into a grid of discrete, non-overlapping cells, representing the trajectory as a sequence of visited cells rather than raw GPS coordinates. An example of a gridded region is shown in Figure \ref{fig:inputoutput}. This discretization of GPS points is necessary for the modeling approach used in this paper. In this study, the entire region is partitioned into uniform square cells represented by the set $G$, where each cell has a unique identifier and a defined geometric shape. The total number of unique cells is denoted by $|G|$. Each GPS-based trajectory is transformed into a grid-based trajectory representation for further analysis. Let $f(.)$ be a function that maps the GPS point to the grid cell enclosing it based on the GPS point coordinates and the geometry of the grid cells. Using this function, the GPS-base trajectory of $T_p$ can be expressed as the ordered sequence of grid cells $T_g=\left\{g_1,g_2,\dots,g_n\right\}$, $g_i=f(p_i)$, $g_i\in G$. Therefore, $g_i$ is the cell encompassing the GPS point $p_i$.

\textbf{Segment-based Trajectory} is a chronological sequence of connected road segments representing the vehicle's path. Here, the segment-based trajectory is denoted by $T_r=\left\{r_1,\dots,r_j,\dots, r_m\right\}$ where $r_j\in R$ and $m$ is the number of connected road segments forming the segment-based trajectory.

\begin{figure*}[thpb]
      \centering
      \includegraphics[scale=0.5]{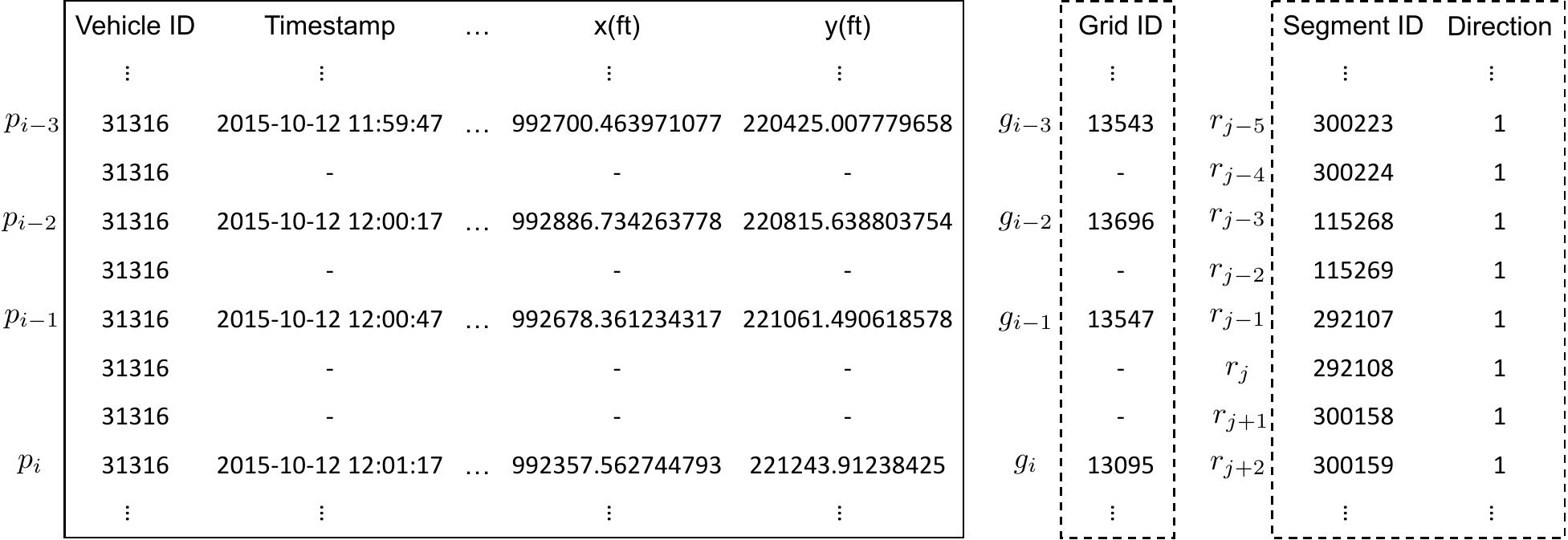}
      \caption{Data structure: The left box shows a snippet of the raw trajectory consisting of $n$ noisy GPS points, denoted as ($p_i|i\in\left\{1,\dots,n\right\}$), sorted by timestamp. The middle box displays the corresponding grid cell IDs for each GPS point, denoted as ($g_i|i\in\left\{1,\dots,n\right\}$). The right box presents the map-matched trajectory, consisting of $m$ road segments, denoted as ($r_j|j\in\left\{1,\dots,m\right\}$).}
      \label{fig:datastructure}
   \end{figure*}
This study introduces an offline map-matching framework that processes a raw, erroneous GPS-based trajectory, through first converting it to grid-based trajectory, and then outputs the most probable path as a segment-based trajectory. Figure \ref{fig:inputoutput} represents an example of the raw input (left), transformed input (middle), and the output (right). Figure \ref{fig:datastructure} presents a snippet of the data structure for a sampled raw GPS trajectory, along with its grid-based and road segment-based representations. Since a low sampling rate in data collection results in a sparse GPS-based trajectory representation, the size of the segment-based trajectory may be equal or greater than that of the GPS-based trajectory, as multiple road segments can exist between two consecutive GPS points. For simplicity, in the following sections, $T_g$, $T_r$ are denoted as $g$ and $r$, respectively.

\section{Methodology}
\subsection{Problem Statement}
The goal of this paper is to create a deep learning model that estimates the actual route taken by a vehicle in the form of a sequence of road segments when provided with a sequence of sparse, raw GPS coordinates. A seq2seq model is a well-suited approach to achieve this objective. This study employs a transformer-based encoder-decoder model to predict the most likely route based on a sequence of spatially unique grid cells onto which the noisy raw coordinates are projected. 

\subsection{Seq2seq Model} A Seq2seq model with neural network \cite{sutskever2014sequence} is an architecture commonly used in natural language processing (NLP) for various tasks such as machine translation, text summarizing, conversation generation, and in other fields that involve sequential data such as image captioning, music generation, and drug discovery \cite{grechishnikova2021transformer}. The basic idea behind seq2seq models is to take a sequence of input tokens and generate a sequence of output tokens corresponding to the input sequence. The most common architecture for seq2seq models is the encoder-decoder. The encoder extracts the important features of the input sequence and produces continuous representations of the input that summarize all relevant information. Then, the decoder recursively generates the output sequence one element at a time based on the decoded context and the previously generated element. The key advantage of seq2seq models is their ability to handle variable-length input and output sequences, which makes them well-suited for tasks where the length of the input and output sequences can vary. Additionally, the neural network-based seq2seq models are capable of learning complex patterns and interdependencies within sequential data, yielding enhanced performance compared to conventional neural network models. RNN and the transformers are the most common model architectures used in the encoder-decoder models. However, the vanishing or exploding gradient problem is a well-known challenge in training RNN encoder-decoder models. This issue arises when the gradients used for backpropagation become extremely small or large, causing the model to fail to converge effectively. In addition to the vanishing or exploding gradient problem, RNN encoder-decoder models are computationally expensive, especially for long sequences. Attention mechanisms mitigate gradient problems in RNN encoder-decoder models, enhancing performance, especially with long sequences, by focusing on specific input parts during output generation. Nevertheless, RNN encoder-decoder models with attention mechanisms remain computationally demanding due to the added computational overhead required for calculating attention weights for each input token.
The transformers are a newer model architecture introduced by \cite{vaswani2017attention} that addresses the limitations of RNN encoder-decoder models. Transformers diverge from RNNs in their architecture, opting instead for stacks of self-attention mechanisms to capture the interconnections within input sequences and their relevance to output sequences. The decoder component leverages parallelized matrix multiplications, simultaneously processing all input sequence elements, as opposed to the sequential nature of RNNs. This parallel processing is a pivotal factor contributing to transformers' enhanced speed and efficiency compared to traditional RNN-based models, which tend to have a more sequential processing approach, often leading to extended training times and slower inference speeds. Additionally, the transformers have been shown to achieve the state-of-the-art performance on a variety of tasks \cite{vaswani2017attention}.

\subsection{Objective Function}
Let $g = (g_1, g_2, ..., g_{n})$ be the transformed version of $T= \left\{ (x_1, y_1), (x_2, y_2), \dots, (x_n, y_n) \right\}$, representing the variable-length input sequence, here in this study, a sequence of grid cells, each of which encompasses each GPS point, and $r = (r_1, r_2, ..., r_{m})$ be the variable-length output sequence, i.e., sequence of road segments, $m$ is the length of the output sequence. The seq2seq model maps $g$ to $r$ by learning a conditional probability distribution over the set of possible output sequences as:

\begin{equation}
\label{eq:1}
p(r|g) = \prod_{t=1}^{m} p(r_t|r_1,\dots,r_{t-1},g)
\end{equation}

The model learns a probability distribution from input-output pairs in training, adjusting parameters to maximize output sequence likelihood given input; the seq2seq model aims to maximize this log-likelihood, described as:

\begin{equation}
\label{eq:2}
\begin{split}
arg \max \limits_{\theta}\ P(r|g) = argmax \log P(r|g) \\= argmax \sum_{t=1}^{m} log P(r_t|g_{n}, r_{<t}).
\end{split}
\end{equation}

The objective can be optimized using backpropagation of cross-entropy loss through time, which computes the gradients of the objective with respect to the model parameters and updates the parameters using gradient descent. Once the model is trained, it can be used to generate segment-based trajectories for new input sequences.

In particular, the transformer \cite{vaswani2017attention} is employed in the seq2seq model to build the surrogate model for mapping $g$ to $r$.

\subsection{The Transformer}
\begin{figure*}[thpb]
      \centering
      \includegraphics[scale=0.65]{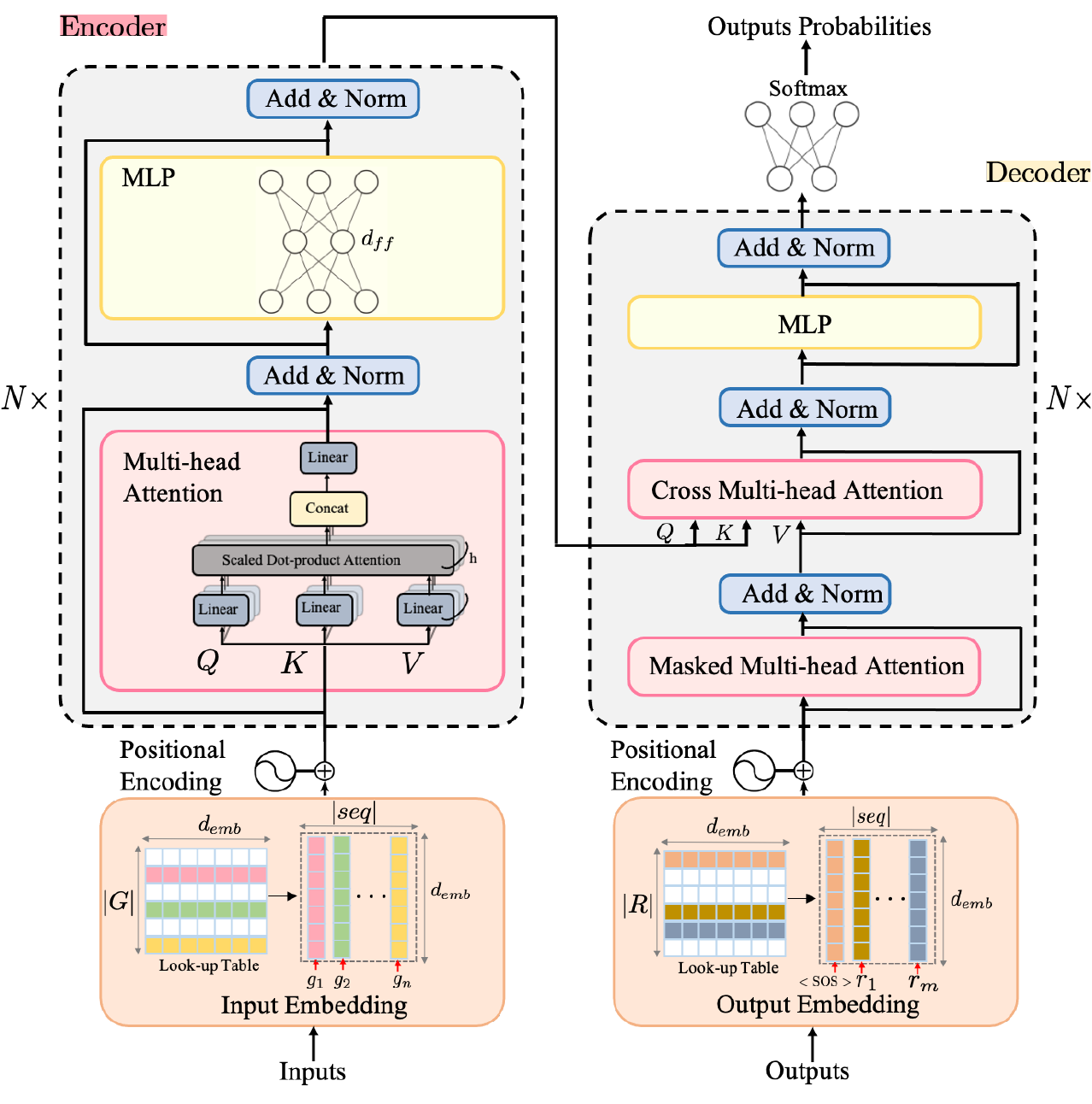}
      \caption{Transformer Model. The encoder on the left and the decoder on the right both contain an embedding layer followed by positional encoding and $N$ transformer blocks. $|G|$ and $|R|$ are the size of unique grids and unique road segments in the region (counterpart of vocabulary size in NLP). $d_{emb}$, $d_{mlp}$, $N$, and $h$ are the model hyperparameters. }
      \label{fig:model}
   \end{figure*}

The neural network architecture with the transformer, depicted in Figure \ref{fig:model}, comprises two primary components: the encoder and the decoder. Detailed explanations of each component are provided in the subsequent sections.
\subsubsection{Encoder}
The encoder in the transformer is composed of an embedding layer, a positional encoding layer, and a stack of $N$ identical layers, each layer composed of a multi-head self-attention mechanism, and a position-wise fully connected feed-forward neural network each of which is followed by a normalization residual connection layer (\textit{Add \& Norm}), each of which is explained in details as follows. 

\textbf{Embedding.} The input to the proposed surrogate map-matching model is a sequence of grid cells encompassing the original GPS points. These grid cells are obtained by partitioning the entire road network into uniform square sections, and the sequence of these grid cells aligns with the sequence of the GPS points (details explained in the preprocessing section). Every grid cell is assigned a unique numerical index, which is then translated to a fixed-size one-hot vector. Consequently, the one-hot vector is typically long but sparse. The size of the one-hot vectors, represented by $|G|$, aligns with the number of distinct cells in the region, including two additional labels to mark the sequence's beginning and the end. The equivalent of $|G|$ in NLP is the source vocabulary size.
The embedding layer consists of a trainable embedding matrix operating similarly to a lookup table that correlates an embedding vector with each distinct grid cell. These embedding vectors are learned during the training process of the neural network.

\textbf{Positional Encoding.} Within this layer, the positional encoding matrix is generated for the input embedding matrix, and subsequently, it is incorporated into the embedded matrix by matrix addition. Given that, unlike recurrent neural networks (RNNs), the transformer architecture lacks the innate ability to capture the order of the elements in the sequence, an additional mechanism is required to infuse positional context into the model. Positional encoding informs the transformer about token positions in a sequence, aiding in distinguishing elements by their sequence order, which is essential for tasks where element arrangement matters.
The positional encoding utilizes a set of fixed sinusoidal functions with different frequencies to represent the position of individual tokens within the sequence. These sinusoidal functions are combined with the token embeddings to generate the final input representation encompassing semantic and positional context. We use the same formula for calculating the positional encoding proposed in \cite{vaswani2017attention}, which is as follows:

\begin{equation}
PE_{(pos,i)} = sin(\frac{pos}{10000^{(\frac{2i}{d_{emb}})}}),
\end{equation}
\begin{equation}
PE_{(pos,2i+1)} = cos(\frac{pos}{10000^{(\frac{2i}{d_{emb}})}}),
\end{equation}
where $pos \in \{1,\dots, n\}$ is the position of the token in the sequence of length $n$, $i \in \{1,\dots, d_{emb}\}$ is the dimension of the positional encoding, $d_{emb}$ is the model's embedding dimension also referred as model dimension. 

\textbf{Multi-head Attention.} Multi-head attention allows the model to focus on different parts of the input sequence simultaneously and learn different aspects of the data in parallel. A multi-head attention layer takes three identical copies of the output of the positional encoding layer or the previous self-attention layer as the query matrix denoted as $(Q)$, the key matrix denoted as $(K)$, and the value matrix denoted as $(V)$. The query matrix represents the elements from the previous layer that are being used to generate attention scores for each element in the current layer. Each row of the query matrix is a query vector corresponding to each sequence element. Similarly, the key matrix represents elements from the previous layer. It consists of key vectors, each corresponding to an element in the sequence, and is used to determine how well other elements match or relate to the corresponding query vectors. Key vectors are crucial in computing attention scores, which indicate the importance of other elements for each query. Similarly, the value matrix represents elements from the previous layer and consists of value vectors that hold the information passed to the next layer. The attention scores calculated using query and key vectors determine how much weight each value vector receives when producing the output. Multi-head attention is a combination of $n$ self-attentions. The most common self-attention involves calculating attention scores between the query and key matrices by taking the dot product of $Q$ and $K^T$ and then scaling the result by the square root of the dimension of the key vectors, $d_k$. Therefore, the core operation of self-attention is scaled dot-product attention. The attention scores are passed through the Softmax function to obtain attention weights. The attention weights are then used to compute a weighted sum of the value matrix $(V)$. This aggregation process captures the relevant information from different elements in the input sequence based on their importance.

Multi-head attention applies linear transformations to map input into $h$ lower-dimensional matrices such that $hd_k=d_{emb}$. The weight matrices of these transformations are learned during training.
Therefore, in the multi-head attention mechanism, the scaled dot-product attention process is repeated $h$ times, in parallel, each time using different linearly transformed versions of $Q$, $K$, and $V$. The outputs from these parallel attention heads are concatenated and linearly transformed again to produce the final multi-head attention output.
The following equations represent the operations in the mathematical format:

\begin{equation}
\text{Attention}(Q,K,V) = \text{Softmax}(\frac{QK^T}{\sqrt{d_{k}}})V,
\end{equation}
\begin{equation}
\text{Multi-head Attention} = \text{Concat}(\text{head}_1,\dots,\text{head}_h)W^o,
\end{equation}
\begin{equation}
\text{head}_i = \text{Attention}(QW_i^Q, KW_i^K,VW_i^V),
\end{equation}
where $W_i^Q$, $W_i^K$, $W_i^V$, and $W^o$ are learnable weight matrices. 

\textbf{Normalization Residual Connection.} Normalization Residual Connection layer (Add \& Norm) is the transformer component that helps stabilize and improve the convergence of the training process. It is an element-wise addition of each sub-layer input and its output, followed by layer normalization \cite{ba2016layer}, and it is used after every sub-layer in the transformer blocks within both the encoder and decoder.
Layer normalization normalizes the values along each dimension of the input tensor. This helps to stabilize the training process and improve the flow of gradients during backpropagation. 

\textbf{Feed Forward Neural Network Layer.} The transformer's feedforward neural network or multilayer perceptron consists of two main linear transformations with a Rectified Linear Unit (ReLU) activation function. The input to the feedforward network is the output from the previous layer. This input is passed through a linear transformation called the point-wise feedforward layer. It uses a weight matrix and a bias term to perform a linear mapping of each position separately and identically to a new intermediate representation, $d_{ff}$. After the first linear transformation, the ReLU activation function is applied to introduce non-linearity to the transformation. The result of the activation function is then passed through another linear transformation, similar to the first step. This transformation maps the intermediate representation to the final output with the model embedding dimension. 

\subsubsection{Decoder}
The decoder in the transformer operates in an autoregressive manner, implying that it generates tokens sequentially, with each step's outcome contingent upon previously generated tokens. This process entails an iterative approach, where the generated token from each step is fed back into the decoder for subsequent steps. Similar to the encoder, the decoder contains an embedding layer, a positional encoding layer, and a stack of $N$ identical layers, each encompasses a masked multi-head layer, a cross multi-head attention layer, and a feed-forward neural network layer. These components are subsequently followed by a normalization residual connection layer, concluding with a final linear layer connected to a Softmax function. In the decoder's self-attention, elements are masked to block access to subsequent elements, allowing the encoder to utilize input and prior output tokens but not tokens from the output sequence that follow the current position. The cross multi-head attention layer allows the model to look at and consider information from every position in the input sequence simultaneously and in turn captures the information in the input sequence that is pertinent or necessary for generating the current token in the output sequence. Indeed, the cross-multi-head attention assists the decoder in establishing alignment between the produced output and the relevant input context. This alignment is achieved by utilizing the previous layer of the decoder for queries ($Q$) allowing the model to consider its own context, while the keys ($K$) and values ($V$) are sourced from the output of the encoder's final layer providing access to the input sequence. Consequently, every key-value pair signifies a distinct position within the input sequence.
\section{Comparative Evaluation with State-of-the-art Methods}
This section presents a comparative evaluation of our proposed model against several state-of-the-art models. The goal of this evaluation is to demonstrate the effectiveness of our model in performing as a surrogate deep learning model for the map matching task. Collecting ground truth data for GPS trajectories is a challenging and resource-intensive task. To the best of authors' knowledge, the publicly available vehicular trajectory datasets with ground truth labels in our region of interest, New York City, are scarce, making it difficult to perform comprehensive evaluations on real trajectory data. To address this limitation, a data augmentation technique is adopted to generate pairs of noisy trajectories and ground truth trajectories. The following section presents a detailed process of data augmentation that enables easy replication of realistic trajectories.
\subsection{Data augmentation}
\begin{figure}[!t]
     \centering
      \includegraphics[scale=0.5]{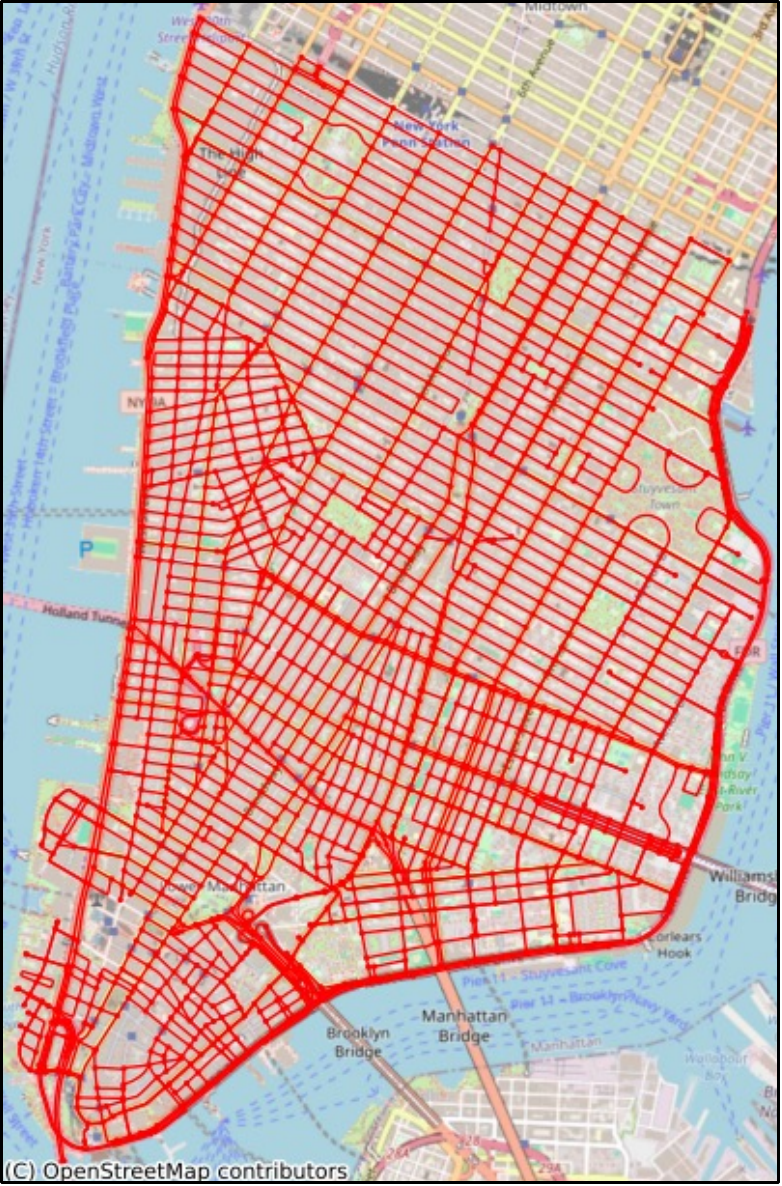}
     \caption{Street network map of Downtown Manhattan, New York. The base map is provided by OpenStreetMap contributors.}
      \label{fig:MN_south}
\end{figure}

\begin{figure}[!t]
     \centering
      \includegraphics[scale=0.45]{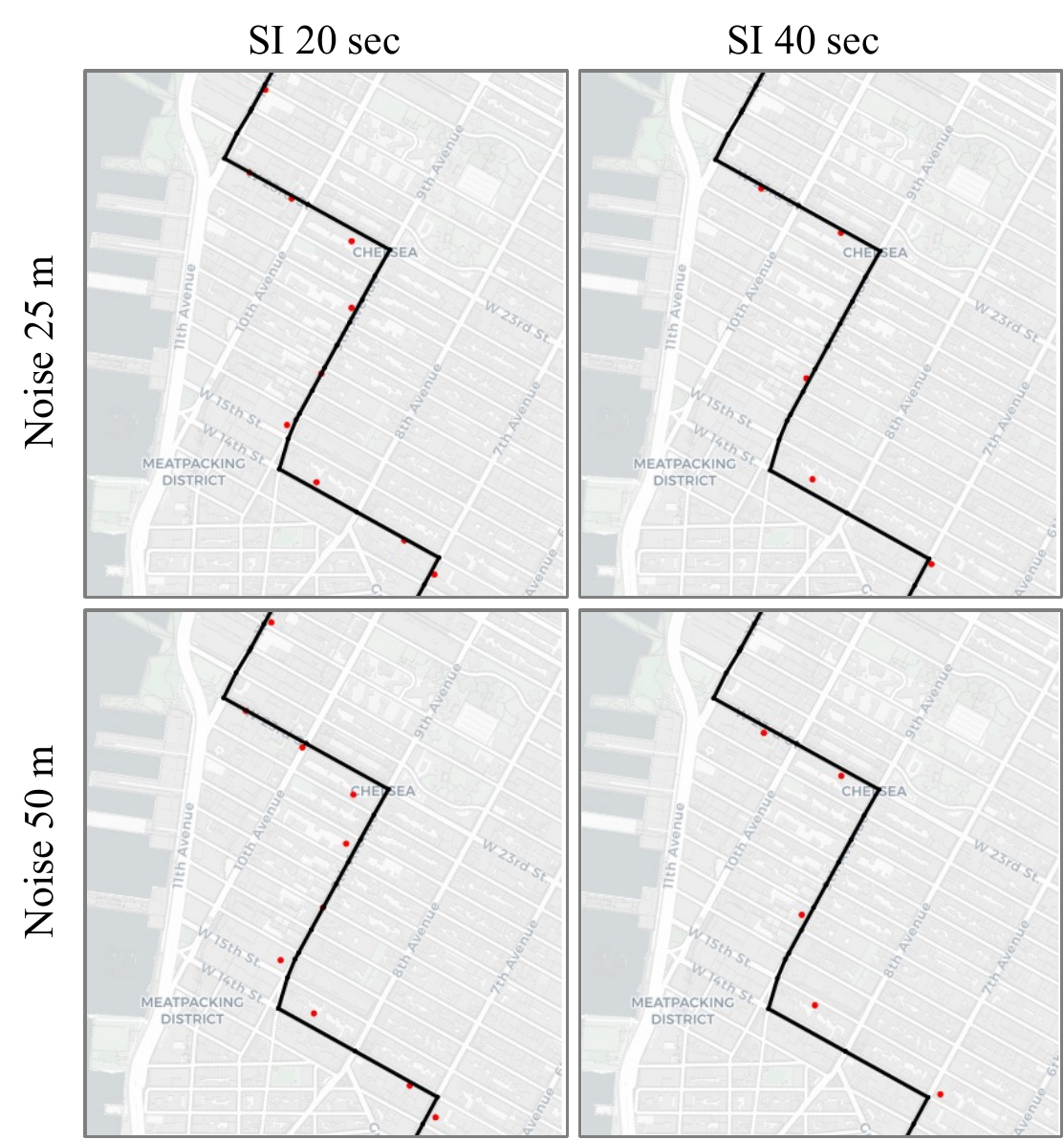}
     \caption{Illustration of an augmented trajectory with different noise level and sampling intervals (SI).}
      \label{fig:noisesettings}
\end{figure}

In this section, four different combinations of noise level and sampling interval are considered in the data generation. Each combination consists of either low noise (25 meters) or high noise (50 meters) with short intervals (20 seconds) or long intervals (40 seconds). Figure \ref{fig:noisesettings} illustrates an example of four different representations of an augmented trajectory with noisy GPS points, each featuring varying noise levels and sampling intervals (SI). For each of the four combinations, 10,000 GPS trajectories are generated for downtown Manhattan road network in New York City shown by Figure \ref{fig:MN_south}. Of these, 80\% (8,000 trajectories) are randomly selected to train the transformer model, and 20\% (2,000 trajectories) are used to test and compare the performances of all three models. Transformer model discards the trajectories with lengths less than four and greater than 200, excluding 16\% of the training and testing data in long interval dataset and around 4\% in the short interval dataset. The test data is identical across all three models to ensure a fair comparison. The synthesized data for the comparison study component of the paper is available to the public on a permanent data repository at \cite{dataset}. The data generation process is as follows.

1. A weighted directed graph representing the structure and geospatial features of the road network is created using OpenStreetMap.

2. To generate more realistic and complex paths beyond the shortest route between origin and destination points, since drivers do not always choose the shortest paths, a perturbation technique is applied. The graph's edge weights are assumed as street lengths except for approximately 30\% of the randomly selected edges, the weights are slightly increased to make these streets less preferable.

3. Origin and destination points for each trajectory are randomly selected from the map. The Dijkstra algorithm is then used to find the shortest path between each pair based on the adjusted edge weights on the weighted directed graph. The path is accepted if its length does not exceed a threshold of 6 miles. This constraint is imposed to avoid overly long routes, simplifying model evaluation in this section. In the actual application during the preprocessing phase, a data segmentation technique is employed to handle long trajectories.

4. The generated routes, structured as sequences of road segments, are considered the true paths from which noisy GPS points are collected. These routes serve as the true labels for the map-matching task.

5. To simulate noisy GPS points, first, vehicle speed data and traversed distances are used to generate true GPS points. It is assumed that the vehicle's speed equals the posted speed limit. Based on the speed and the length of each road segment, true GPS points, representing the vehicle's actual location, are sampled at specified sampling intervals. Then, Gaussian noise with a specified standard deviation is added to each true GPS point to create the noisy GPS points as the input for the map-matching model. 
\subsection{State-of-the-art map-matching methods}
Different map-matching models necessitate various input features to execute map-matching. Some models operate solely on noisy GPS points \cite{newson2009hidden}, while others require additional heading information \cite{quddus2015shortest} or in-vehicle speed information \cite{alrassy2020map}. Due to the absence of speed and heading information in our dataset, and considering that multiple heading changes typically occur within 30-60 second sampling intervals \cite{greenfeld2002matching}—making heading computation non-trivial without comprehensive contextual data—we have chosen to focus on models that perform map-matching using only sequences of noisy GPS points for comparative analysis.

\subsubsection{ST-matching \cite{lou2009map}} ST-matching is one of the popular map-matching algorithms that has gained extensive attention \cite{yang2018fast}. This algorithm takes into account the spatial geometric and topological structures of the road network, as well as the temporal constraints of the trajectories. The parameters for this algorithm are set as follows: the maximum number of candidates is $k = 10$, the search radius is $r = 500$ meters, and the GPS error is $\sigma = 25$ meters for low-noise datasets and $\sigma = 50$ meters for high-noise datasets.
\subsubsection{Leuven-matching: HMM with non-emitting states \cite{meert2018hmm}}
In conventional Hidden Markov Models (HMMs), each state transition absorbs an observation, leading to issues when measurements are less frequent than segment transitions. A novel map-matching algorithm inspired by bioinformatics is introduced by \cite{meert2018hmm} that employs non-emitting states enabling dynamic interpolation without needing an observation for each transition. The parameters of this algorithm are set as follows: the observation noise is $\sigma = 25$ meters for low-noise datasets and $\sigma = 50$ meters for high-noise datasets, possible candidate states per observation 10, and minimum normalized probability of observations as 0.0001.
\subsubsection{Transformer-based surrogate map-matching} The configuration of the porposed transformer-based model is set as follows. The embedding dimension ($d_{emb}$) is set to 512, and the MLP intermediate dimension ($d_{ff}$) is set to 1024. We use $N=4$ transformer blocks for both the encoder and decoder, and each multi-head attention block employs $h=4$ linear transformations, resulting in $d_k=128$. The Adam optimizer with a learning rate of 0.0001 and the cross-entropy loss function is utilized. The batch size is set to 32, and the model trained for a total of 200 epochs. The inference time for map-matching using the constructed transformer model varies with trajectory length, with an average inference time of 0.30s and a standard deviation of 0.16s for trajectories ranging from 6 to 204 road segments.
\subsection{Data structure}
The ST-Matching and Leuven-Matching models both are instantiated with a weighted directed graph representing the road network. In this graph, the edges correspond to road segments, and the nodes correspond to the endpoints of these segments. This graph is created using the OSMnx Python package, which reads and represents street features from OpenStreetMap as a graph \cite{boeing2017osmnx}. These models take as input a sequence of noisy GPS points and output a sequence of road segments, represented by the nodes of endpoints of these segments.

For the transformer model, the same graph generated from OpenStreetMap is utilized. Additionally, a list of geometries is pre-generated to represent the polygons of uniform grid cells, partitioning the region of interest into exclusive equal blocks, each with a unique identifier. More detailed information is provided in Section \ref{subsec:preprocessing}. The transformer model is instantiated with these geometries and two vocabulary objects for input and output: the set of all input tokens (in this spatial context, tokens are the grid cells, each assigned a unique index) and the collection of all output tokens (here the road segments from OpenStreetMap, each assigned a unique index), respectively. The input to the transformer model is initially noisy GPS points, which are mapped to the grid cells within which each point lies, and the output is a sequence of the road segments (endpoints of the segments).
\subsection{Evaluation Metrics}
The performance of each model is evaluated using three different metrics, accuracy, Jackkard similarity, and Bilingual Evaluation Understudy (BLEU) \cite{papineni2002bleu}, mathematically described below:
\begin{equation}
\text{Accuracy} = \frac{1}{N}\sum_{i=1}^{N}\frac{|\hat{T}_i\cap T_i|}{|T_i|}
\end{equation}
where $T_i$ and $\hat{T}_i$ are the $i$th ground truth path and the estimated path in the whole trajectory set $T$. $|.|$ represents the segment count of the trajectory.

The Jaccard similarity, shown by equation \ref{eq:9} is a measure of similarity between two sets. Here, we use this metric to compare each trajectory with model estimation. It is defined as the size of the intersection of the two routes composed of road segments divided by the size of their union and yields a value between 0 and 1, where 0 indicates no similarity (no common elements) and 1 indicates complete similarity (identical sets). 
\begin{equation}
\label{eq:9}
\text{Jackkard} = J(T_i,\hat{T}_i) = \frac{1}{N}\sum_{i=1}^{N}\frac{|\hat{T}_i\cap T_i|}{|\hat{T}_i\cup T_i|}  
\end{equation}

BLEU is a metric commonly used to evaluate the quality of machine-generated text, particularly in the context of machine translation. However, in the context of this paper, BLEU is utilized to evaluate the similarity of the predicted segment-based trajectory with the ground truth trajectory. Similar to the text evaluation, here it is assumed to operate by comparing n-grams (contiguous sequences of n road segments) in the candidate and reference trajectories. The intuition behind BLEU is that a good mapping should have similar n-gram statistics to those of the ground truth trajectories. BLEU which is shown in equation \ref{eq:12}, has two parts, \textit{precision}, see equation \ref{eq:10}, and \textit{brevity penalty}, see equation \ref{eq:11}. Precision which counts how many contiguous sequences of ``n" items (n-grams) in the candidate route match those in the reference route and divides this count by the total number of n-grams in the candidate routes and \textit{brevity penalty} that penalizes the short mappings \cite{papineni2002bleu}. 

\begin{equation}
\label{eq:10}
Precision(\hat{T}_i, T_i) = \frac{\text{count of n-gram matches between }\hat{T}_i, T_i}{\text{Count of n-grams in } \hat{T}_i},
\end{equation}
\begin{equation}
\label{eq:11}
BP = \begin{cases} 
      1 & |\hat{T}_i| > |T_i| \\
      \exp{(1-\frac{|\hat{T}_i|}{|T_i|})} & |\hat{T}_i| \leq |T_i|, \\
   \end{cases},
\end{equation}
\begin{equation}
\label{eq:12}
BLEU = BP \exp(\frac{1}{n}\sum_{i=1}^n\log(Precision_i))
\end{equation}
where $n$ is the maximum n-gram length considered. In this paper, we report the equally weighted average of n-grams for n-values ranging from 1 to 4.
\subsection{Comparison results}
\begin{figure}[!t]
     \centering
      \includegraphics[scale=0.48]{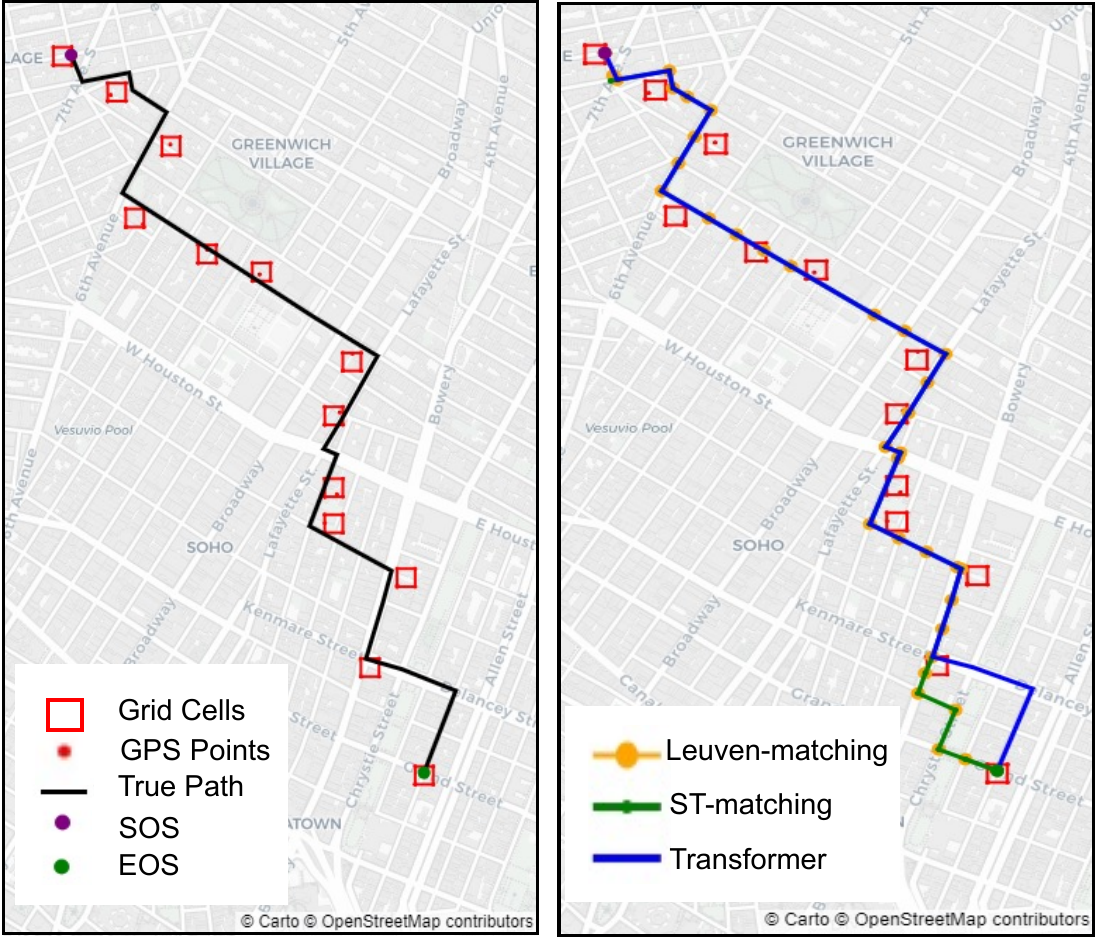} 
     \caption{An example of trajectory map-matching results (right figure) using ST-Matching (green), and Leuven-Matching (orange), Transformer (blue) for an augmented trajectory with the true path represented in black (left figure). Between two initial GPS points, the transformer accurately estimates the true path compared to the other two methods.}
      \label{fig:MM_synth_ex1}
\end{figure}

The map-matching results are presented in table \ref{tab:comparison}. Although increasing the noise and the sampling interval degrades the performance of all three methods, the transformer consistently outperforms the other two state-of-the-art methods, ST-Matching and Leuven-Matching, across all four settings. One possible reason for this observation is that the state-of-the-art methods rely on the shortest path as the most probable route between two points. In contrast, the deep learning model not only learns the noise patterns but also captures drivers' driving patterns and route selection behavior by analyzing large collections of trajectories. An example of a successful capture of the true path using the transformer model is illustrated in Figure \ref{fig:MM_synth_ex1}. While the need for the large training dataset is a limitation of deep learning models, techniques such as transfer learning can address issues related to sparse datasets. Although this is not the focus of this study, training the model with a sparse dataset is an area for future research.   
\begin{table}[ht]
\centering
\caption{Comparison of Models Across Different Settings}
\label{tab:comparison}
\begin{threeparttable}
\begin{tabular}{llccc}
\toprule
\textbf{Setting} & \textbf{Model} & \textbf{Accuracy} & \textbf{J} & \textbf{BLEU} \\
\midrule
\multirow{3}{*}{LN-SI\tnote{1}} & Transformer & 0.8529 & 0.9182 & 0.8451\\ 
                           & ST-matching & 0.7741 & 0.8531 & 0.7991\\
                           & Leuven-matching & 0.7462 & 0.7978 & 0.7053\\
\midrule
\multirow{3}{*}{LN-LI\tnote{2}} & Transformer & 0.8004 & 0.8728 & 0.8012\\
                           & ST-matching & 0.6985 & 0.7931 & 0.7296\\
                           & Leuven-matching & 0.5972 & 0.6524 & 0.5608\\
\midrule
\multirow{3}{*}{HN-SI\tnote{3}} & Transformer & 0.8386 & 0.8984 & 0.8343\\
                           & ST-matching & 0.7170 & 0.8326 & 0.7239\\
                           & Leuven-matching & 0.8158 & 0.9025 & 0.7862\\
\midrule
\multirow{3}{*}{HN-LI\tnote{4}} & Transformer & 0.7698 & 0.8473 & 0.7725\\
                           & ST-matching & 0.6632 & 0.7743 & 0.6915\\
                           & Leuven-matching & 0.5987 & 0.6734 & 0.5776\\
\bottomrule
\end{tabular}
\begin{tablenotes}
\footnotesize
\item[1] LN-SI: Low Noise - Short Interval
\item[2] LN-LI: Low Noise - Long Interval
\item[3] HN-SI: High Noise - Short Interval
\item[4] HN-LI: High Noise - Long Interval
\end{tablenotes}
\end{threeparttable}
\end{table}
\section{Experiment with real data}
\label{sec:exp}
\subsection{Baseline}
\subsubsection{HMM} The goal of this paper is to introduce robust surrogate modeling for map-matching. In particular, in this paper, we are building a transformer-based map-matching model to operate as a surrogate model for our in-house cost-based HMM map-matching engine. This engine incorporates an innovative map-matching algorithm that merges speed data from On-Board Diagnostics (OBD-II) sensor modules within the vehicles with the GPS trajectory data. This fusion yields a notable enhancement in robustness and accuracy, and the algorithm overperformed a reliable off-the-shelf map-matching platform, resulting in an accuracy of 97.45\% \cite{alrassy2021obd}. 
\subsubsection{RNN-based encoder-decoder with attention mechanism}
We also compare the transformer-based map-matching model with the RNN-based encoder-decoder model. The RNN-based encoder-decoder model and its variants have been adopted by \cite{ren2021mtrajrec,feng2020deepmm}. Feng et al. trained and evaluated DeepMM, the LSTM-based encoder-decoder with attention mechanism, with the trajectory data generated and recorded in part of Beijing containing 4,058 road segments, and the sampling
interval of 60 seconds. They achieved the highest accuracy of 66\% \cite{feng2020deepmm}. Similarly, Ren et al. trained and evaluated MTrajRec, the GRU-based encoder-decoder with attention mechanism and multi-task learning component, with the trajectory data recorded in Jinan, Shandong with 2,571 road segments in the area, and they achieved the highest recall accuracy of 74.98\% for the data with a high sampling rate and 69.72\% for the data with a low-sampling rate \cite{ren2021mtrajrec}. In another research work, Liu et al. trained a graph-based encoder-decoder model with 64k trajectories collected in the northeast of Beijing with 8.5K road segments and achieved an accuracy of 66.22\% for trajectories with 30-second time intervals. They have also shown that their model outperformed MTrajRec \cite{liu2023graphmm}.

The main components of the RNN-based encoder-decoder architecture are two layers: the embedding layer and the RNN layer. Therefore, the main difference between the RNN-based encoder-decoder with the transformer-based encoder-decoder is its recurrent layer versus the self-attention layer in the transformer. The architecture of the RNN-based encoder-decoder model can be found in \cite{cho2014learning}.
In an RNN-based encoder-decoder, during the encoding, the RNN layer processes all the tokens on the input sequence iteratively and reutilizes the hidden state of an iteration as the input for generating the hidden state for the next element. The last hidden state of the RNN is passed as input to the first iteration of the decoder. Therefore, when processing a sequence of inputs, the RNN layer generates a sequence of hidden states, with each hidden state being a function of the current input and the previous hidden state. In particular, a bidirectional Gated Recurrent Unit (Bi-GRU) in the RNN layer of the encoder, and a regular GRU in the RNN layer of the decoder are utilized. The structure of the Gated Recurrent Unit (GRU) is presented by \cite{chung2014empirical}.

GRU captures the sequential information in the data by maintaining an internal hidden state that evolves over time. It has a gating mechanism that controls the flow of information within the network, making it capable of learning and remembering longer sequences compared to traditional RNNs. The gating mechanism consists of an update gate and a reset gate, which control how much of the previous hidden state should be retained and how much new information should be incorporated in each time step.
A Bi-GRU is a variant of the GRU architecture that processes input sequences in both forward and backward directions. The input sequence is fed into two separate GRU layers, one processing the input sequence in the forward direction and the other processing the input sequence in the backward direction. The outputs of the two layers at each time step are concatenated and passed to the next layer or used as the final output. 

\subsection{Dataset}
\begin{figure}[thpb]
      \centering
      \includegraphics[width=0.4\textwidth]{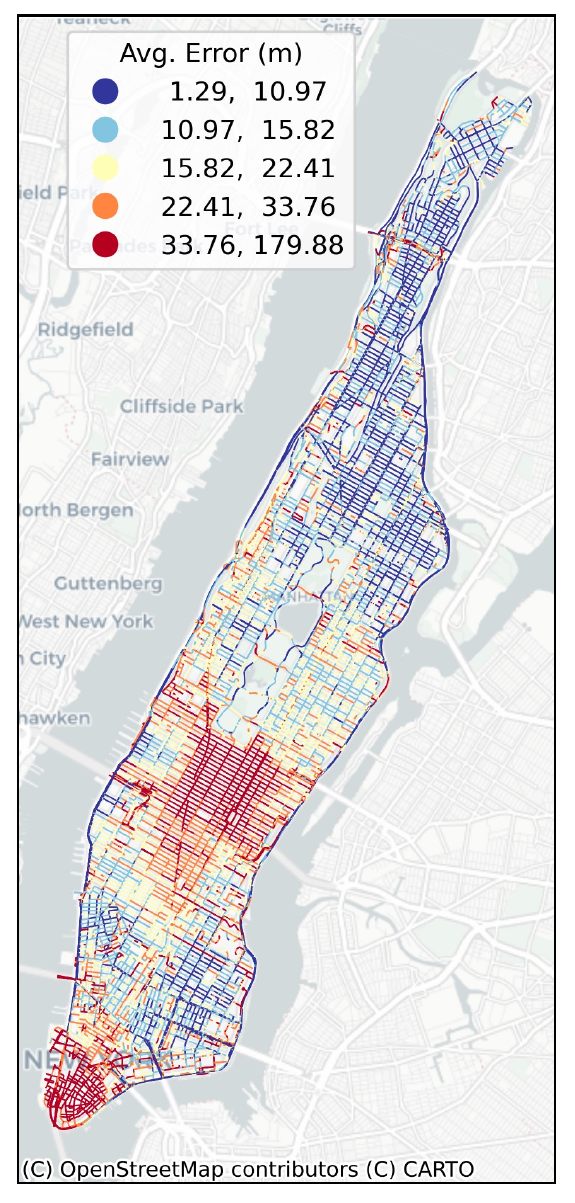}
      \caption{Spatial distribution of GPS error (in meters) aggregated per road segment.}
      \label{fig:error_map}
   \end{figure}

To train our neural network model, we employ an 18-month dataset comprising 8,854,325 telematic data points associated with 32,097 trajectories recorded in Manhattan, New York City spanning from January 2015 to June 2016. This dataset is sourced from approximately 4,500 city-owned vehicles, each equipped with telematics devices and managed by the New York City Department of Citywide Administrative Services (NYC DCAS). The telematics devices recorded vehicle positions at roughly 30-second intervals. This data was initially map-matched using an in-house map-matching engine, developed by Alrassy et al. \cite{alrassy2021obd}. Trajectories processed through this map-matching engine accounted as the labels for our study. The GPS error statistics including the mean, median and standard deviation of the GPS errors are 15.73m (51.6ft), 6.95m (22.8ft), and 23.13m (75.9ft), respectively. Additionally, the spatial distribution of the average GPS error per road segment is illustrated in Figure \ref{fig:error_map}. As expected, the error has a spatially heterogeneous pattern and the highest error is observed in midtown and downtown of Manhattan due to the existence of tall towers. For the digital map, LION Single Line Street Base map spatial data (version 17D), which is publicly available from the NYC Department of City Planning, is utilized.  
   
\subsection{Pre-processing}
\label{subsec:preprocessing}
\begin{figure}[thpb]
      \centering
      \includegraphics[scale=0.6]{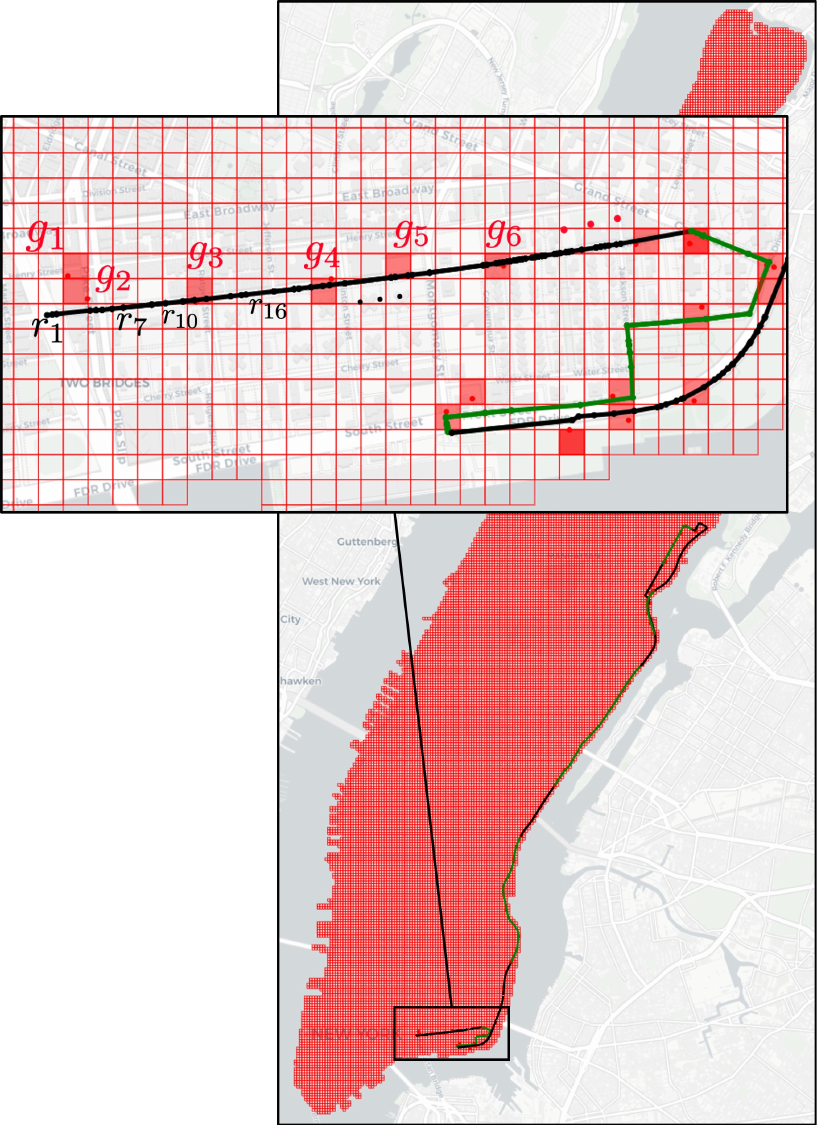}
      \caption{Trajectory and Grid Representations. For illustration purpose, black and green colors are used to depict the trajectory segments, alternating between the two colors in every other segment.}
      \label{fig:grepr}
   \end{figure}

To integrate trajectory data into the encoder-decoder model, a series of preprocessing steps are required. 
The first step is transforming the raw point-based trajectories into a grid-based trajectories. The concept of using a grid representation for maps has its origin in the field of object localization in robotics \cite{milstein2008occupancy}. It has also found practical applications in space-free vehicle trajectory recovery \cite{park2018sequence,wang2019deep}. Furthermore, it became a valuable strategy in seq2seq map-matching models, facilitating the conversion of the continuous input domain into a discrete domain \cite{feng2020deepmm,ren2021mtrajrec,jiang2023l2mm,liu2023graphmm}. In this procedure, each individual data point was mapped onto non-overlapping square cells, effectively partitioning the entire region into uniform grid cells (Figure \ref{fig:grepr}). During this process, various grid cell sizes were explored. It was observed that employing an excessively small grid size resulted in an extensive number of grid cells, potentially compromising the performance of the seq2seq model. Consequently, a range of grid sizes between 100-200 was tested. The grid size of 150ft was considered the optimal choice, yielding better accuracy. Subsequently, rather than using a sequence of points, each input trajectory is now depicted as a sequence of grid cells, while on the other hand, the output trajectory is a sequence of road segments. Due to the low sampling rate, usually, the length of the input trajectory (the count of grid cells) is smaller than the length of the corresponding output trajectory, i.e., the size of the road segment sequence representing the complete route. As a result of the partitioning, the input set yielded a total of 25,146 distinct grids, while the output set comprised 17,965 unique segments.

The second preprocessing step focuses on segmenting lengthy trajectories for training. Encoder-decoder models, constrained by memory limitations, impose a maximum sequence length. When dealing with extensive input sequences, including lengthy trajectories, they often surpass this limit. To resolve this challenge, we split long input trajectories into shorter segments that conform to the model's constraints. This segmentation addresses the memory issue and also alleviates computational complexity. We have set the maximum trajectory length to 100 and 50 road segments for transformer- and RNN-encoder-decoder model, respectively, and apply a short overlapping between fragments when splitting the trajectories. This strategy helps us comply to the length constraint while with overlapping approach preserves the correlations between the individual trajectory segments. The smaller maximum length is selected for the Bi-GRU encoder-decoder since training the RNN models for long sequences is challenging.
Based on the trajectory data and the statistics, it's observed that, on average, there are approximately 4 road segments between every two consecutive GPS points. To ensure that the maximum length of the road segment sequence does not exceed the limit, we have chosen to set the maximum length of the GPS sequence to 20 GPS points for training the transformer and 8 GPS points for training the RNN model. Example of trajectory segmentation is shown in Figure \ref{fig:grepr}. For illustration purpose, we use black and green colors to depict the trajectory sections, alternating between the two colors in every other section.

\subsection{Implementation Detail}
After evaluating different sets, model hyperparameters are configured as follows. The embedding dimension ($d_{emb}$) is set to 512, and the MLP intermediate dimension ($d_{ff}$) is set to 2048. We use 8 transformer blocks for both the encoder and decoder, and each multi-head attention block employs 16 linear transformations, resulting in $d_k=32$. During training, we utilize the Adam optimizer with a learning rate of 0.0001 and the cross-entropy loss function. The batch size is fixed at 200, and the model undergoes training for a total of 32 epochs.
For the RNN model, the embedding dimension is set to 1024. We employ the Adam optimizer with a learning rate of 0.0005 for both the encoder and decoder. Similarly, the batch size is fixed at 200, and the model undergoes training for a total of 32 epochs reaching to the point that the loss function converges.
We implemented the model in Python using the PyTorch library and trained it on a machine equipped with NVIDIA RTX 6000 GPUs.

\section{Results}
\begin{figure}[!t]
     \centering
      \includegraphics[scale=0.33]{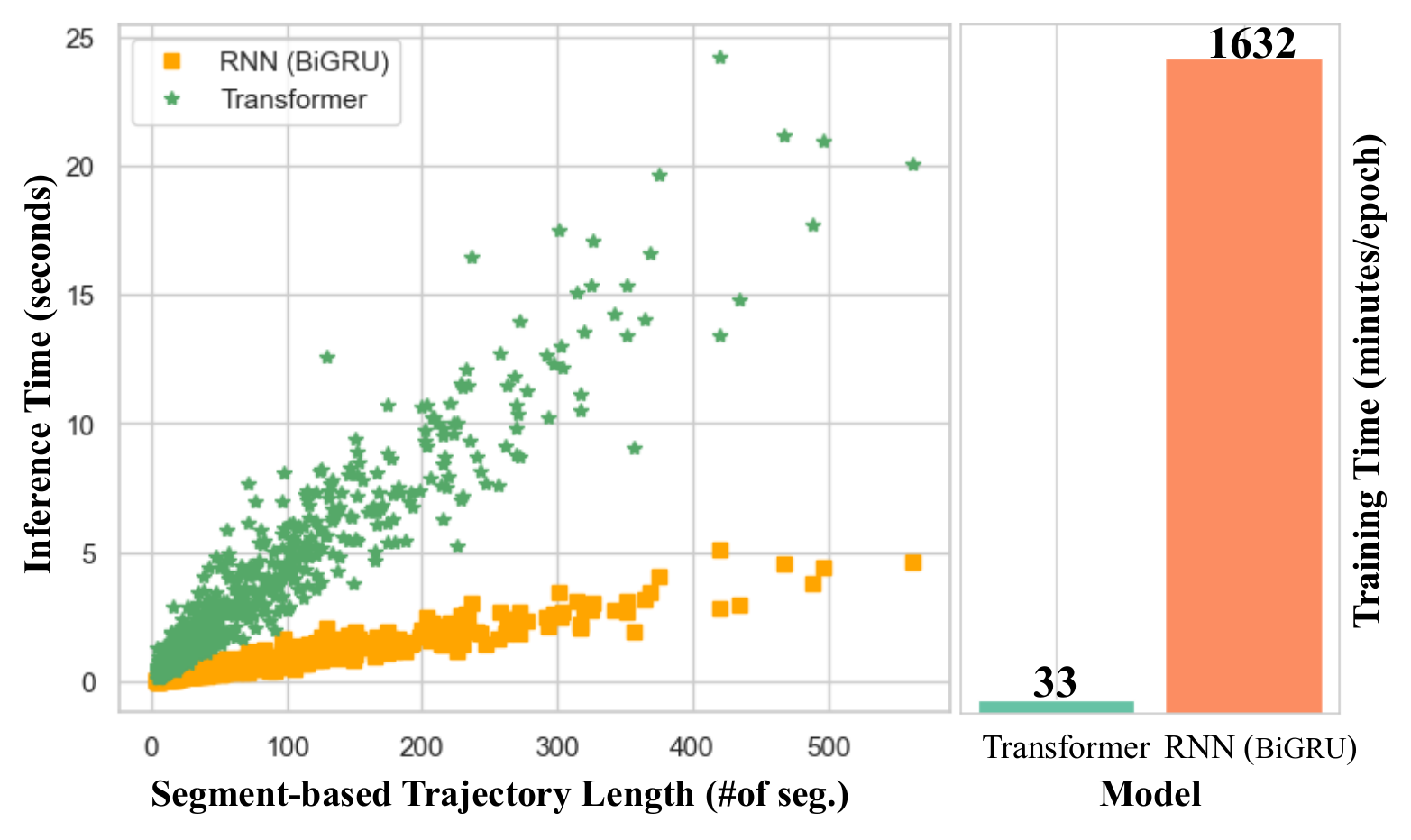}
     \caption{Training and Inference Time.}
      \label{fig:timing}
\end{figure}

The trajectory data is randomly split into 98\% (31,455 trajectories) for training and 2\% (642 trajectories) for testing. The complexity of the Manhattan road network with around 17,965 unique segments, posed a significant challenge given our relatively sparse dataset of 31,455 trajectories. To ensure comprehensive coverage of the road network and effective learning from this data, we opted for a split of 98\% for train and 2\% for test (resulting in 642 unseen trajectories for testing the model). This decision was made to maximize the model's exposure to the data covering the complex road network of Manhattan, which was crucial for training. However, despite this high training proportion, our model demonstrated robust performance on the unseen data, indicating that it did not overfit and retained its generalization capability. The average length of trajectories used for testing the model's performance is 16 kilometers (10 miles). After trajectory segmentation to adhere to the constraint of the maximum length, the data size expands to 272,722 to train the transformer model and 402,361 to train the Bi-GRU encoder-decoder model. The models are trained using the train data set and then evaluated with the test data set. The transformer encoder-decoder model outperformed the Bi-GRU encoder-decoder and resulted in an accuracy of 75.2\%, Jackkard similarity (\textit{J}) of 78.1\%, and BLEU of 75.6\%. While, the Bi-GRU encoder-decoder model resulted in the accuracy of 58.3\%, Jackkard similarity (\textit{J})  of 53.7\% and BLEU of 53.9\%. Results are summarized in table \ref{tab:Performance}. The transformer model reached convergence after 32 epochs, with each epoch completing in 33 minutes. On the other hand, the RNN-based model achieved convergence in 10 epochs, although each epoch took a longer 1632 minutes. However, a closer examination of the inference times for various trajectory lengths, as depicted in Figure \ref{fig:timing}, reveals that the RNN-based encoder-decoder model exhibits faster inference times than the constructed transformer model. This variance can be attributed to the distinctive architectural characteristics of the models.
Generally, the transformers-based encoder-decoder model exhibits more effective and faster convergence compared to the RNN-based counterpart. Several factors contribute to this phenomenon. First, the self-attention layer's sequential operation complexity is $O(1)$ while the recurrent layers' sequential operation is $O(n)$. The self-attention layer captures relationships and dependencies between different parts of the input sequence. Then, the number of operations required to connect all positions (or elements) within the sequence remains constant, regardless of the sequence length. This is typically achieved by using matrix operations that can be efficiently paralleled, making self-attention well-suited for capturing long-range dependencies in sequences. A recurrent layer, on the other hand, involves operations that depend on the previous states of the layer. Thus, the computation at each step is dependent on the computation of the previous step. As a result, the number of sequential operations required by a recurrent layer scales linearly with the length of the sequence. This is represented by $O(n)$, where $n$ is the length of the sequence. Second, the complexity per layer of the self-attention layer used in the transformer is $O(n^2d)$ where $d$ is the model dimension (in our case the model embedding dimension). In a self-attention layer, attention scores are computed between all pairs of positions in the input sequence, resulting in a quadratic complexity with respect to the sequence length $n$. Additionally, each attention score involves a dot-product computation between two $d$-dimensional vectors, which contributes to the $d$ factor in the complexity. While the recurrent layer's complexity is $O(nd^2)$. At each position, there's a multiplication of a $d$ dimensional vector (the input) with a $d$ dimensional matrix (the recurrent weight matrix), leading to the $d^2$ factor in the complexity expression and the overall complexity of $O(nd^2)$ Since in our study, the model dimension is larger than the input sequence, thus, the recurrent layer is more complex than the self-attention layer.

\begin{figure}[!t]
     \centering
      \includegraphics[scale=0.5]{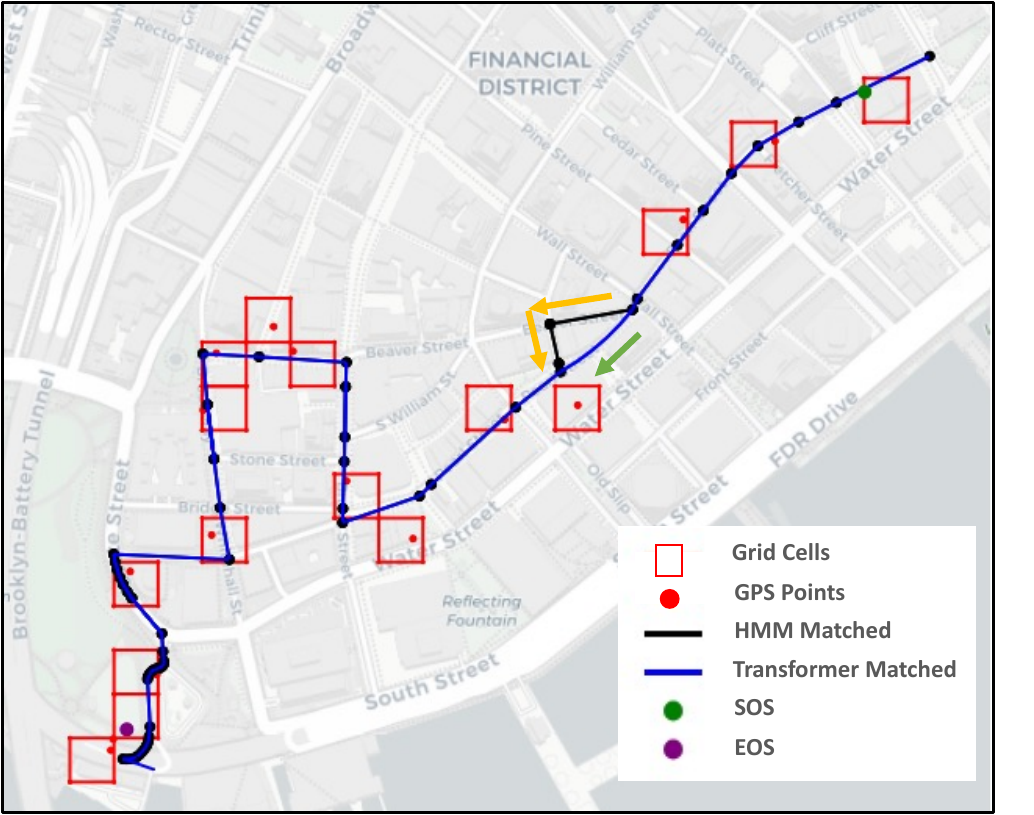}
     \caption{An example of map-matched results using a Transformer-based model (in blue) compared to a cost-based Hidden Markov Model (HMM) map-matching algorithm (in black) for a trajectory from Pearl Street to Battery Park. The Transformer-based model, represented by the blue line, exhibits a more direct and streamlined path, capturing the mobility pattern based on extensive historical data. In contrast, the Cost-based HMM map-matching algorithm, depicted by the black line, suggests a peculiar detour along Beaver Street in the mid-way of the journey. This deviation from the straight path is a characteristic of the HMM's probabilistic nature, where it may make decisions based on local observations without global context.}
      \label{MM_ex1}
\end{figure}

\begin{figure*}[thpb]
     \centering
      \includegraphics[scale=0.5]{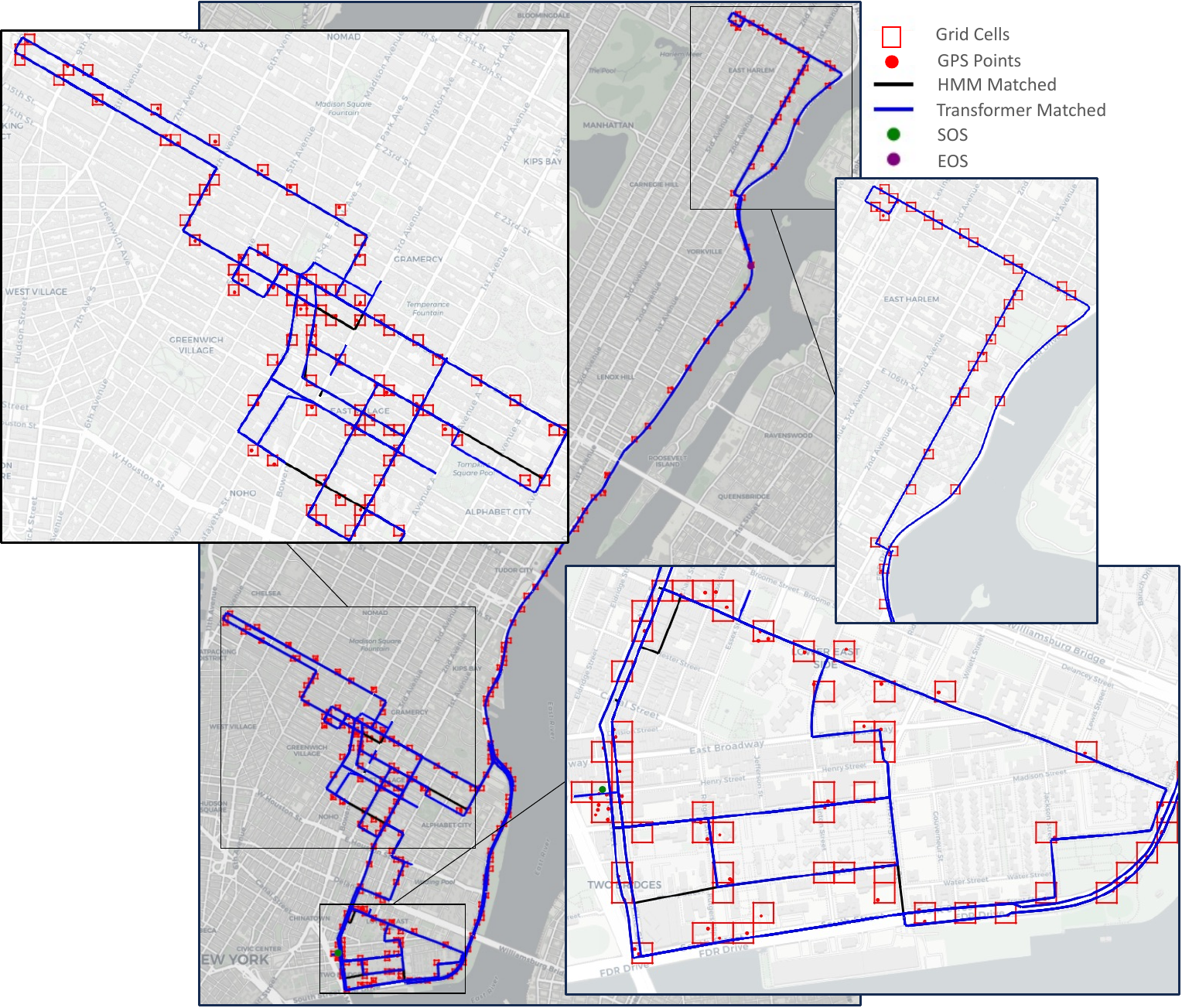}
     \caption{Example of a lengthy trajectory that originated at the intersection of Henry Street and Pike Street and concluded at the intersection of East End Ave and FDR Dr. The outcome reveals a substantial alignment between the route matched with the map utilizing a cost-based HMM and the route matched with a transformer-based approach.}
      \label{fig:MM_ex2}
\end{figure*}

Overall, the transformer model has demonstrated promising performance in the map-matching task. However, it should be noted that the deep transformer model benefits significantly from leveraging large trajectory data during training. This enables it to capture the mobility patterns from processing a wealth of trajectories, ultimately leading to its superior performance. Figure \ref{MM_ex1} presents an example of map-matched results using a transformer-based model (in blue) compared to a cost-based Hidden Markov Model (HMM) map-matching algorithm (in black) for a trajectory from Pearl Street to Battery Park. The Transformer-based model, represented by the blue line, exhibits a more direct and streamlined path, capturing the mobility pattern based on extensive historical data. In contrast, the Cost-based HMM map-matching algorithm, depicted by the black line, suggests a peculiar detour along Beaver Street in the mid-way of the journey. This deviation from the straight path is a characteristic of the HMM's probabilistic nature, where it may make decisions based on local observations without global context.
Figure \ref{fig:MM_ex2} illustrates another example of a lengthy trajectory that originated at the intersection of Henry Street and Pike Street and concluded at the intersection of East End Ave and FDR Dr. The outcome reveals a substantial alignment between the route matched with the map utilizing a cost-based HMM and the route matched with a transformer-based approach. However, there are a few noticeable discrepancies between the two methods. For instance, in Allen Street, the transformer recommends continuing on a straight path, while the HMM algorithm suggests making a turn onto Hester Street. The former alignment corresponds more closely to the collected GPS points and aligns better with the intuition.

In the map-matching results, additional discrepancies are present, which degrades the quality of the matching process. These discrepancies stem from the manner in which inferences associated with segmented routes are combined. In this study, we employed a brute-force approach for merging the inferences of two adjacent segments with partial overlap; the model retains the first segment and combines the remaining portion of the second segment. This approach represents one of the limitations of our work, and addressing this issue calls for a more rigorous solution.

\begin{table}[!t]
    \centering
    \caption{Models' Performance Metrics}
    \label{tab:Performance}
    \begin{tabular}{lccc}
    \toprule
         \textbf{Model} & $\textbf{Accuracy}$ & \textbf{J} & \textbf{BLEU} \\
         \midrule
         Transformer&0.752 &0.781 &0.756\\
         RNN(BiGRU)& 0.583  &0.537 &0.539\\
         \bottomrule
    \end{tabular}
    
\end{table}


\section{Conclusion}
This study introduces a context-aware deep-learning-based trajectory map-matching framework that predicts the most likely navigated path as an ordered sequence of road segments from noisy and sparse GPS data. Traditional methods rely on geometric proximity, topological features, and simplified assumptions, such as uniform noise distribution and shortest-path heuristics for route estimation. These approaches often use probabilistic models or handcrafted rules, which may not generalize well across diverse urban environments. Unlike conventional methods, large-scale neural networks can infer contextual factors such as non-uniform noise patterns, driver trajectory behavior, and complex road network structures from real-world data. By leveraging large-scale trajectory datasets, these models capture spatial uncertainties and mobility patterns, improving the accuracy and robustness of trajectory map-matching.

Building on the success of large language models, large-scale neural network-based foundation models are now accelerating various scientific domains. Examples include Aurora, trained on large-scale weather and climate data \cite{bodnar2024aurora}, and ChemBERTa, designed for molecular data to aid drug discovery \cite{chithrananda2020chemberta}. Recent studies have also envisioned a shift toward context-aware trajectory-powered foundation models \cite{choudhury2024towards,musleh2022towards} for accelerating downstream mobility tasks.

In line with this vision, this study presents an NLP-enabled map-matching framework by reframing map-matching as a sequence-to-sequence translation task, similar to those in NLP. The model adopts a transformer-based encoder-decoder architecture, widely used in natural language processing. By tokenizing noisy GPS trajectories and leveraging self-attention mechanisms, the model learns context-aware representations of driver trajectory behavior, road network structures, and environmental factors that contribute to noise variations. This approach enables the model to effectively translate raw GPS points into their most likely navigated routes, capturing spatial dependencies and contextual information beyond traditional heuristic-based methods. This approach aligns with the trajectory-as-language perspective presented by \cite{musleh2024let}, which seeks to replace traditional, rule-based trajectory analysis with learning-based sequence models that generalize across different urban environments. Similar to how large language models capture latent linguistic structures without predefined rules, this method allows the model to learn spatial correlations, mobility preferences, and GPS noise distributions directly from trajectory data.

The effectiveness of the developed framework is demonstrated through evaluation on real-world GPS traces in Manhattan, New York. The model improved trajectory map-matching accuracy by leveraging both local road network features and global contextual dependencies, serving as a practical step toward trajectory-powered foundation models.

For deep learning-based map-matching frameworks successful out-of-domain generalization requires accounting for variability in road network, built environment, and local driving behaviors. Pretraining on multi-city trajectory datasets and transfer learning \cite{raffel2020exploring} can further enhance the model’s ability to generalize to trajectory analysis in different city’s road networks while preserving core spatial and temporal dependencies that drive human mobility. Expanding the training data to include cities with diverse network structures, mobility patterns, and GPS noise distributions could transform this model into a scalable, region-agnostic trajectory analysis tool, contributing to the broader vision of a universal trajectory foundation model for urban mobility \cite{choudhury2024towards}.

Scaling this model to large cities and multi-city applications requires addressing the increasing complexity of road networks, particularly the growth in unique grid locations ($G$) and road segments ($R$). This scaling challenge is analogous to vocabulary expansion in NLP models, where handling larger vocabularies necessitates techniques such as adaptive embeddings \cite{baevski2018adaptive,joulin2017efficient}, Mixture of Softmax Experts (MoE) \cite{shazeer2017outrageously}, language-clustered vocabularies \cite{chung2020improving}, and hierarchical location encoding \cite{ainslie2020etc,park2023pre}. Further investigation is needed to understand the relationship between road network size and model performance, particularly in exploring whether similar NLP strategies can enhance trajectory analysis in large-scale road networks. While NLP models provide a strong conceptual foundation, their direct adaptation to trajectory modeling may introduce unique challenges. A deeper exploration within the context of trajectory data is essential for developing scalable, multi-city trajectory models capable of generalizing across diverse urban landscapes.

Future research in NLP-enabled geospatial trajectory analysis is needed to explore the feasibility and effectiveness of these strategies in managing large-scale road networks. While multilingual NLP models provide a valuable conceptual foundation, their direct adaptation to trajectory modeling introduces unique challenges, such as spatial dependencies, network topology constraints, and variations in real-world mobility patterns. Investigating region-aware tokenization, hierarchical location encoding, and adaptive attention mechanisms in the context of trajectory data will be critical to developing scalable, multi-city trajectory models capable of generalizing across diverse urban landscapes.

\section{Acknowledgments}
The authors acknowledge the support from Google and the Tides Foundation under the grant ``EMS Resource Deployment Modeling'' and the Columbia University Urban Technology Pilot Award. This work was also partially supported by the Center for Smart Streetscapes, an NSF Engineering Research Center, under grant agreement EEC-2133516. Authors also acknowledge Dr. Patrick AlRassy for his help with the original HMM algorithm data processing.

\bibliographystyle{IEEEtran}
\bibliography{IEEEexample}

\vfill

\end{document}